%% file: acl_latex.tex
\documentclass[11pt]{article}

\usepackage[final]{acl}

\usepackage[]{linguex}
\usepackage{times}
\usepackage{latexsym}
\usepackage{CJKutf8}
\usepackage{amsmath}
\usepackage{expex}
\usepackage{multirow}
\usepackage{longtable}
\usepackage{array}
\usepackage{booktabs}
\usepackage{tabularx}

\usepackage[dvipsnames]{xcolor}

\usepackage[T1]{fontenc}

\usepackage[utf8]{inputenc}

\usepackage{microtype}

\usepackage{inconsolata}

\usepackage{graphicx}

\title{How Much Do LLMs Know About Chinese Zero Pronouns?}

\author{Yifei Li, Guanyi Chen\thanks{Corresponding Author}, {\normalfont and} Tingting He\\
  Hubei Provincial Key Laboratory of Artificial Intelligence and Smart Learning, \\
  National Language Resources Monitoring and Research Center for Network Media, \\
  School of Computer Science, Central China Normal University \\
  \texttt{muzifeitu@mails.ccnu.edu.cn, \{g.chen, tthe\}@ccnu.edu.cn}}

\begin{document}
\begin{CJK}{UTF8}{gbsn}

\maketitle
\begin{abstract}
Zero Pronouns (ZPs) are a pervasive linguistic phenomenon in pro-drop languages such as Chinese and have long posed a challenge for natural language processing systems. Although Large Language Models (LLMs) perform well on many Chinese language tasks, their ability to process ZPs remains poorly understood. We conduct a systematic investigation of LLMs’ handling of Chinese ZPs through a sequence of linguistically motivated tasks, including identification, referentiality classification, referential type classification, resolution, and translation. A diverse set of LLMs is evaluated across all tasks. Our results show that Chinese ZPs remain highly challenging for current LLMs, particularly for upstream tasks such as identification and referentiality classification. Performance on downstream tasks, such as ZP translation, is also consistently low: even state-of-the-art reasoning-oriented LLMs correctly translate fewer than half of Chinese ZPs into English.
\end{abstract}

\section{Introduction}

Languages such as Chinese, Japanese, and Korean allow the frequent pragmatic omission of overt pronouns, a phenomenon commonly referred to as zero pronouns \citep[ZP,][]{huang1984distribution}. Based on an analysis of a large Chinese–English dialogue parallel corpus, \citet{wang2018translating} reported that approximately 26\% of pronouns overtly expressed in English are not explicitly realised in Chinese. This pervasive use of zero pronouns has led linguists to characterise Chinese as a \emph{``cool''} language \citep{huang1984distribution,huang1989pro} or as \emph{discourse-oriented} \citep{cao1979functional}, highlighting its strong reliance on discourse context for meaning recovery.

As an illustration, consider the Chinese question ``你今天看见比尔了吗?'' (Did you see Bill today?). In response, a speaker may produce several reduced utterances that convey the same meaning as the fully explicit sentence ``我看见他了'' (Yes, I saw him). For example, one may say ``$\emptyset$看见他了'' (\emph{Yes, $\emptyset$ saw him}), ``我看见$\emptyset$了'' (\emph{Yes, I saw $\emptyset$}), or even ``$\emptyset$看见$\emptyset$了'' (\emph{Yes, $\emptyset$ saw $\emptyset$}). In these examples, the symbol $\emptyset$ marks positions where pronouns are omitted but can be readily inferred from the discourse context.

With the advent of large language models (LLMs), these models have demonstrated strong reasoning capabilities and advanced language proficiency~\citep{mao-etal-2024-gpteval,jones2025large}. Focusing on Chinese, \citet{liu2024systematic} systematically investigated LLMs’ ability to process a wide range of Chinese-specific linguistic phenomena, including ba-constructions, classifiers, and ellipsis, among others. Nonetheless, \emph{the extent to which LLMs can handle zero pronouns—one of the most salient phenomena in Chinese—remains largely underexplored.}

Before the age of LLMs, computational studies of ZP primarily focused on ZP resolution, in which classification models were built to select the antecedent of a ZP from a set of candidate mentions~\citep{chen-ng-2013-chinese,chen2014chinese,yin-etal-2017-chinese,yin-etal-2018-deep,yin-etal-2018-zero,song-etal-2020-zpr2,konno-etal-2021-pseudo}. Another major line of research investigated downstream tasks, most notably the translation of zero pronouns~\citep{xu-etal-2022-guofeng,wang-etal-2023-survey}. In contrast, comparatively little attention has been paid to upstream tasks, such as identifying the positions of ZPs~\citep{wang-etal-2019-one} or determining whether a ZP is anaphoric. These tasks are, intuitively, indispensable components in the processing pipeline for a comprehensive understanding of ZPs.

\begin{figure*}[t]
    \centering
    \includegraphics[width=0.95\linewidth]{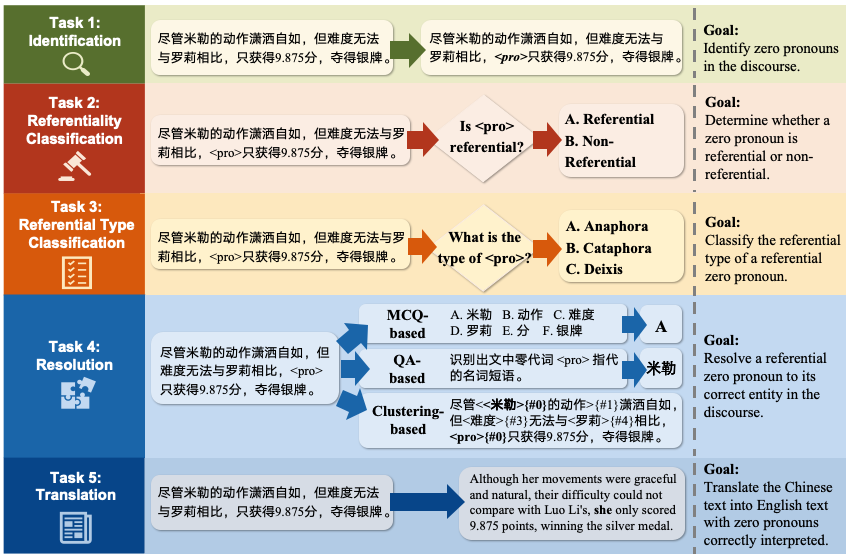}
    \caption{An overview of our tasks. The translation of the example: \emph{Although Miller’s movements were fluid and confident, the difficulty could not compare with Li Luo’s. She received a score of 9.875 and won the silver medal.}}
    \label{fig:placeholder}
\end{figure*}

To conduct a comprehensive and systematic examination of LLMs’ ability to process zero pronouns in Chinese, this work considers a range of tasks defined over the OntoNotes dataset. Specifically, we examine not only \emph{ZP Resolution}, but also its downstream task, namely \emph{ZP Translation}, as well as several upstream tasks, including \emph{ZP Identification} (i.e., determining whether a position contains a zero pronoun), \emph{Referentiality Classification} (i.e., deciding whether a zero pronoun is referential), and \emph{Referential Type Classification} (i.e., determining whether a zero pronoun is an anaphora or a cataphora).

For each of these tasks, we proceed as follows. (1) We consider one or more task formulations. Taking ZP resolution as an example, we include multiple-choice-question–based, question-answering–based, and clustering-based resolution settings. (2) We design task-specific evaluation protocols tailored to Chinese, taking into account that Chinese is a character-based language. (3) We evaluate a broad spectrum of LLMs, ranging from small models (e.g., 4B parameters) to very large models (e.g., 671B parameters), and from base models (e.g., Qwen and GPT-4.1) to reasoning-oriented models (e.g., DeepSeek-R1 and GPT-o4).

In this study, we aim to address the following two research questions:
\begin{description}
    \item[$\mathcal{RQ}_1$] To what extent can LLMs perform a range of Chinese Zero Pronoun Centred tasks, including Identification, Referentiality Classification, Referential Type Classification, Resolution, and Translation?
    \item[$\mathcal{RQ}_2$] How are LLMs’ abilities across these Zero Pronoun tasks related? In particular, do models that perform well on downstream tasks (e.g., translation) also perform well on upstream tasks (e.g., identification and referentiality classification)?
\end{description}






\section{Methodology}

Unlike prior ZP-centred tasks—which typically either provide the ZP position and ask models to resolve it, or ask models to translate sentences from pro-drop languages into English—our investigation is designed as a stack of sequential tasks to test whether an LLM can \emph{truly and comprehensively} understand ZPs in Chinese (see Figure 1 for an overview). Intuitively, given a discourse, before resolving the ZPs it contains, a human (and thus an LLM) must first determine whether and where zero pronouns occur (i.e., ZP Identification). Because ZPs are not always anaphoric in context, the system must then decide whether an identified ZP is referential (i.e., Referentiality Classification).

Moreover, while most prior work on pronoun resolution emphasises anaphora, other referential configurations, such as cataphora and deixis, also occur and are referential. It is therefore important to determine the referential type of a referential ZP (i.e., Referential Type Classification), so that a resolution system knows where to search for its antecedent. Ultimately, if an LLM truly understands a ZP, it should be able to both resolve it and translate it appropriately.

It is also worth noting that, based on prior work on annotating ZPs in pro-drop languages, the tasks considered in this study (especially Identification and Referentiality Classification) are challenging not only for native speakers but even for trained linguistic experts. Successfully tackling these tasks is likely to require LLMs to exhibit strong discourse-level reasoning capabilities.

In what follows, we describe each task in our investigation along with its corresponding evaluation protocol. We conduct our study using the OntoNotes dataset, and the preprocessing steps undertaken to adapt the dataset to our task settings are detailed in Section~\ref{sec:data}.

\subsection{ZP Identification}

A single Chinese discourse may contain multiple ZPs. For example, the sentence introduced earlier, ``$\emptyset$看见$\emptyset$了'' contains two ZPs within a single clause. In extreme cases, a discourse may include more than 100 zero pronouns. Accordingly, we define the \textbf{ZP Identification} task as follows: given a discourse, predict the positions of all ZPs.

\paragraph{Evaluation.}
ZP identification is evaluated using precision, recall, and F1 score, computed by comparing the predicted ZP positions with the gold-standard reference positions.

\subsection{Referentiality Classification}

A ZP may be non-referential, a typical case being when it corresponds to an arbitrary noun phrase (NP). For example, in the instruction ``$\emptyset$进入房间前要脱鞋'' (\emph{before entering the room, $\emptyset$ must take off shoes}), the dropped NP denotes people in general rather than a specific referent.

By contrast, a ZP is referential if its antecedent appears in the surrounding discourse or if it is deictic. A deictic NP is a referring expression that lacks a textual antecedent but instead refers to a speech participant or the situational context. For instance, in the sentence ``$\emptyset$要走了'' (\emph{$\emptyset$ am leaving}), the ZP refers to the speaker.

Formally, for each ZP, the \textbf{Referentiality Classification} task requires an LLM to determine whether the ZP is referential or non-referential. We deliberately avoid formulating this task as anaphoricity classification—that is, classifying whether a ZP is anaphoric—because deictic NPs, although referential, are not anaphoric. Nevertheless, such deictic ZPs are relevant to the subsequent resolution and translation stages and therefore need to be retained at this step.

\paragraph{Evaluation.} As a standard binary classification task, referentiality classification is evaluated using Precision, Recall, and F1.

\subsection{Referential Type Classification}

Given a referential ZP, we define \textbf{Referential Type Classification} as the task of determining whether the ZP is (1) anaphoric, i.e., its antecedent appears in the preceding discourse; (2) cataphoric, i.e., its antecedent appears in the subsequent discourse; or (3) deictic, i.e., it is not anchored to any textual antecedent.

\paragraph{Evaluation.} Referential type classification is evaluated using Precision, Recall and F1.

\subsection{ZP Resolution}

In prior work, pronoun resolution has been formulated and addressed in many different ways~\citep{mao2024survey}, depending on factors such as the input assumptions (e.g., whether the system is provided with candidate mentions) and the target phenomena (e.g., whether the system aims to resolve anaphora or co-reference, or to perform mention linking). In this study, we try the following three task settings.

\paragraph{MCQ-based Resolution.} In the context of ZP resolution, most prior studies have focused on anaphoric ZPs and formulated the task as a classification problem: given a predefined set of candidate antecedents, a model is trained to select the most plausible one \citep{chen-ng-2013-chinese,chen2014chinese,yin-etal-2017-chinese,yin-etal-2018-deep,yin-etal-2018-zero}. When adapting this formulation to LLMs, the task naturally becomes a Multiple-Choice Question (MCQ), where the candidate antecedents are provided as answer options. Accordingly, the \textbf{MCQ-based Resolution} task requires an LLM to select the correct antecedents for a referential ZP given its surrounding context. Performance is evaluated using accuracy over the MCQ responses. 

\paragraph{QA-based Resolution.} \citet{song-etal-2020-zpr2} were among the first to formulate ZP resolution in an end-to-end manner, training a model to directly identify the antecedent of a ZP from context without being provided with candidate antecedents. We adopt a similar setting by defining a \textbf{Question-Answering (QA) based Resolution} task for LLMs: given a discourse and a marked ZP, the LLM is asked to answer the question ``\emph{What does $\langle$pro$\rangle$ refer to?}'', where $\langle$pro$\rangle$ denotes the ZP. Performance is evaluated using accuracy, determined by whether the antecedent produced by the LLM is in the same reference chain as the target ZP.

\paragraph{Clustering-based Resolution.} Entity-mention models in coreference resolution typically formulate resolution as a clustering task~\citep{cardie-wagstaff-1999-noun}. Under this paradigm, all mentions in a discourse are first grouped into clusters, and the mentions that belong to the same cluster as a target pronoun are treated as its antecedents. \citet{le2023large} were the first to introduce this idea into LLM-based coreference resolution, and we adopt the same paradigm for \textbf{Clustering-based ZP resolution}. As illustrated in Figure 1, mentions in the input text are first marked with special tokens indicating spans to be annotated (e.g., 罗莉 $\to$ [罗莉](\#)). The LLM is then instructed to assign a cluster identifier to each marked span (e.g., [罗莉](\#) $\to$ [罗莉](\#cluster\_1)). Notably, this task formulation naturally allows LLMs to handle deictic zero pronouns. For example, a deictic ZP that refers to a speech participant can be assigned to a cluster consisting of all ZPs that refer to the same participant, even in the absence of an explicit textual antecedent.

As for evaluation, classic resolution metrics such as MUC~\citep{vilain-etal-1995-model} or B$^3$~\citep{bagga1998algorithms} are not well-suited to our setting. Although the LLM clusters all mentions, our primary interest lies in the resolution of zero pronouns. Accordingly, we extract the mentions from the cluster to which a target ZP is assigned and treat them as a bag\footnote{NB: Different from `set', a bag (also called a multiset) is a collection where duplicates are allowed. The intersection of two bags keeps each element as many times as it appears in both bags.} of mentions, denoted by $R$. This bag is then compared against the gold-standard reference chain containing the same ZP, which is likewise represented as a bag of mentions and denoted by $G$. 

Because Chinese is a character-based language, we further define the bags of all characters corresponding to $R$ and $G$ as $R_c$ and $G_c$, respectively. Precision and recall are then defined as:
$$\text{Precision} = \frac{|R_c \cap G_c|}{|R_c|}, \text{Recall}=\frac{|R_c \cap G_c|}{|G_c|}.$$
The F1 score is computed accordingly.

\subsection{ZP Translation}

Similar to previous benchmarks on \textbf{ZP translation} \citep{xu-etal-2022-guofeng}, the ZP Translation task requires an LLM to translate a Chinese discourse in which ZPs are not explicitly marked, and evaluates whether these ZPs are correctly and appropriately rendered in the resulting English translation.

\paragraph{Evaluation.} Given the substantial body of work demonstrating that LLMs achieve strong overall performance in Chinese–English translation (e.g., \citet{huang2024evaluating}), we do not evaluate general translation quality in this study. Instead, we focus exclusively on the accuracy of ZP translation.

For this purpose, \citet{xu-etal-2022-guofeng} proposed the AZPT metric, which computes ZP translation accuracy by automatically aligning each ZP in Chinese with its translated pronoun in English. However, after manually inspecting its outputs, we found that AZPT exhibits low validity. This limitation likely stems from the metric’s underlying assumption that a Chinese ZP should always be translated into an English pronoun. In practice, this assumption often fails, as ZPs may be more appropriately translated as full NPs or proper names in English. Moreover, automatic alignment tools such as GIZA++ struggle to reliably handle such alignments. 

We also experimented with evaluating ZP translation accuracy using an LLM-as-a-judge paradigm. However, we found that determining whether a ZP has been correctly translated is itself a highly challenging task, even for LLMs, making them unreliable as automatic judges in this setting.

Therefore, we ultimately resorted to a human evaluation for zero-pronoun translation. Specifically, we sampled 120 sentences from the dataset that contain at least one ZP. For each sample, the target sentence—together with its immediately preceding and following sentences—was evaluated. Each item was presented with its surrounding context and the corresponding English translation produced by each LLM under evaluation, and the positions of zero pronouns in the Chinese sentences were explicitly marked.

We recruited two annotators who are native speakers of Chinese and fluent in English. For each instance, the annotators were asked to judge whether the ZPs in the Chinese source were correctly translated in the English output. The inter-annotator agreement, which Cohen's Kappa measures, is 0.73. In cases of disagreement, the annotators discussed the instance and reached a consensus. The final ZP translation accuracy was then computed based on these adjudicated annotations. 

\section{Experiment Setup}

In this section, we describe the dataset and the models used in this study.

\subsection{Data} \label{sec:data}

We use the Chinese portion of the OntoNotes 5.0 dataset\footnote{\url{http://catalog.ldc.upenn.edu/LDC2013T19}}
, which was released for the CoNLL-2012 shared task and has been widely adopted in prior research on ZP resolution. To our knowledge, it is the only dataset that provides rich annotations of zero pronouns in Chinese. \footnote{There is also the BaiduZhidao dataset \citep{zhang2019neural}; however, it does not annotate complete reference chains, which makes it unsuitable for several of the tasks considered in this study.} Specifically, for each ZP, the dataset annotates whether it is referential, and for each referential ZP, it further annotates the complete reference chain. The OntoNotes documents are drawn from six genres: Broadcast News (BN), Newswire (NW), Broadcast Conversation (BC), Web Blog (WB), Telephone Conversation (TC), and Magazine (MZ).

We derive the labels for each task primarily from the reference-chain annotations provided in OntoNotes. To reduce computational cost, we do not run our experiments on the full OntoNotes dataset. Instead, we randomly sample 10\% of the documents from each of the six genres, resulting in a total of 172 documents containing 3,607 ZPs. Detailed descriptions of how labels are extracted from reference chains for each task, as well as summary statistics of the resulting dataset, are provided in Appendix~\ref{sec:app_process} and Appendix~\ref{sec:app_statistics}, respectively.

\paragraph{Issue of Data Leakage.} One might argue that, because OntoNotes is a well-known NLP dataset, using it to evaluate LLMs could raise concerns about data contamination~\citep{balloccu-etal-2024-leak}. Nonetheless, we contend that this issue is unlikely to affect our study. In OntoNotes, each document and its reference-chain annotations are stored in separate files, and there has been no prior work that evaluates LLMs on Chinese zero pronouns in a way that would expose models to task-specific supervision during training. Consequently, there is no realistic pathway for an LLM to have been trained on data that directly reveals the information required to solve the tasks considered here. Moreover, the consistently poor performance of LLMs observed in our experiments (see Section~\ref{sec:result}) provides empirical support for this claim.

\subsection{Models}

\input{tab/result}

We include a broad range of LLMs in this study. To examine how model size affects the ability to handle Chinese ZPs, we evaluate Qwen3-4B, Qwen3-8B, and Qwen3-235B~\citep{yang2025qwen3}. We also consider two strong general-purpose LLMs, namely DeepSeek-V3.2~\citep{liu2025deepseek} and GPT-4.1-mini.\footnote{We also experimented with LLaMA2-7B and LLaMA3-8B, but found them to be highly unstable in understanding the task setup and to frequently fail at following instructions. This behaviour is likely attributable to their relatively weak performance in handling Chinese.}

To investigate whether reasoning capability influences performance on Chinese ZP tasks, we additionally include two reasoning-oriented LLMs, DeepSeek-R1~\citep{guo2025deepseek} and GPT-o4-mini~\citep{hurst2024gpt}.\footnote{We use the mini versions of the GPT models to reduce cost; notably, GPT-o4-mini is already approximately ten times more expensive than DeepSeek-R1.}

\section{Results} \label{sec:result}

Table~\ref{tab:result} summarises the main results of our study. The precision and recall values corresponding to all reported F1 scores are provided in Appendix~\ref{sec:appendix_pr}, and performance broken down by genre can be found in Appendix~\ref{sec:appendix_genre}.

\paragraph{General Performance.} To compare the LLMs’ overall performance across all tasks, we compute the Borda count. Specifically, for each task, models are ranked according to the corresponding evaluation metric, and a model ranked $r$ among $K$ models receives a score of $K-r$. The final Borda score for each model is obtained by summing these scores across all tasks. 

Overall, understanding ZPs in Chinese remains highly challenging for LLMs. Across all tasks, models typically achieve performance in the range of 20–40\%, which is substantially lower than that reported for most other NLP tasks~\citep{mao-etal-2024-gpteval}. The results further indicate that larger models tend to handle zero pronouns more effectively, as evidenced by the comparison among the three Qwen models of different sizes. In addition, successful ZP understanding appears to require strong reasoning capability: GPT-o4-mini performed the best among all evaluated models.

\paragraph{ZP Identification and Referential Type Classification.} LLMs’ performance on ZP identification and referential type classification follows the same general trend observed in the overall results: larger and reasoning-oriented models consistently perform better. Among all tasks, ZP identification appears to be particularly challenging. Smaller LLMs (e.g., Qwen3-4B) achieve F1 scores of below 1\%, indicating severe difficulty in detecting ZPs. An illustrative Chain-of-Thought (CoT) example demonstrating how explicit reasoning can aid ZP identification is provided in Appendix~\ref{sec:appendix_reasoning_example}.

\paragraph{Referentiality Classification.} Referentiality classification is also among the most challenging tasks. Despite being a binary classification problem, all LLMs except GPT-o4-mini tend to label every ZP as referential (see the per-class results in Appendix~\ref{sec:appendix_class}). Consequently, while these models achieve F1 scores of around 70\% on referential instances (with recall close to 99\%), their F1 scores on non-referential instances fall below 10\%. In contrast, GPT-o4-mini performs substantially better on the non-referential ZPs. Nonetheless, unfortunately, GPT does not expose the full reasoning chain.

\paragraph{ZP Resolution.} LLMs achieve only 20–30\% accuracy/F1 across the three resolution tasks. Somewhat surprisingly, they perform worst on the MCQ-based resolution task, with most models exhibiting similarly poor results. This weakness can largely be attributed to their very low performance on the BC and TC genres, both of which consist of conversational data (see the per-genre results in Appendix~\ref{sec:appendix_genre}).

A closer inspection reveals that conversational texts contain many coreference chains that are dominated by pronouns and zero pronouns. In such cases, LLMs exhibit a strong bias toward selecting candidate antecedents that are proper names or full noun phrases, even when these options actually refer to different entities, leading to systematic errors in MCQ-based resolution. \footnote{Recall that in the MCQ setting, LLMs are allowed to select multiple options, and a prediction is considered correct only if the chosen set is a subset of the gold reference answer.} 

Nonetheless, a similar issue does not arise in QA-based resolution. This suggests that the bias observed in the MCQ setting may be caused by the confusion introduced by a large set of candidate options. Under the QA formulation, however, LLMs sometimes return phrases that are not referring expressions at all, which keeps overall performance at a relatively low level.

Clustering-based resolution is the most challenging of the three resolution settings, as it requires LLMs to correctly link all mentions within a coreference chain. Consequently, models exhibit very low recall, as shown in Appendix~\ref{sec:appendix_pr}. Larger and reasoning-oriented LLMs tend to achieve higher precision, but reasoning-oriented models are often overly cautious in linking mentions. As a result, DeepSeek-V3.2, which better balances precision and recall, achieves the strongest overall performance on this task. In contrast, GPT-4.1-mini performs poorly because it links very few mentions, yielding a recall of only 2.48\%.

\paragraph{ZP Translation.} The 120 sampled items contain 269 ZPs, yet LLMs successfully translate fewer than half of them. This poor performance is consistent with the findings of \citet{xu-etal-2022-guofeng} on neural machine translation systems from four years ago. More importantly, the differences among the evaluated LLMs are negligible. The only significant difference appears when comparing DeepSeek-V3.2 and DeepSeek-R1. In other words, neither larger model capacity nor stronger reasoning ability appears to substantially improve LLMs’ performance on translating ZPs in Chinese.

\section{Analysis}

In this section, we further analyse the ZP translation task, and its relationship with upstream tasks. We also take a closer look at referentiality and referential type classification, and examine how context size may influence LLMs’ performance on each task. 

\paragraph{How well can LLMs translate ZPs when they are given more explicit information about ZPs?}

\input{tab/translation}
\input{tab/translation_example}

\begin{figure*}
    \centering
    \includegraphics[scale=0.4]{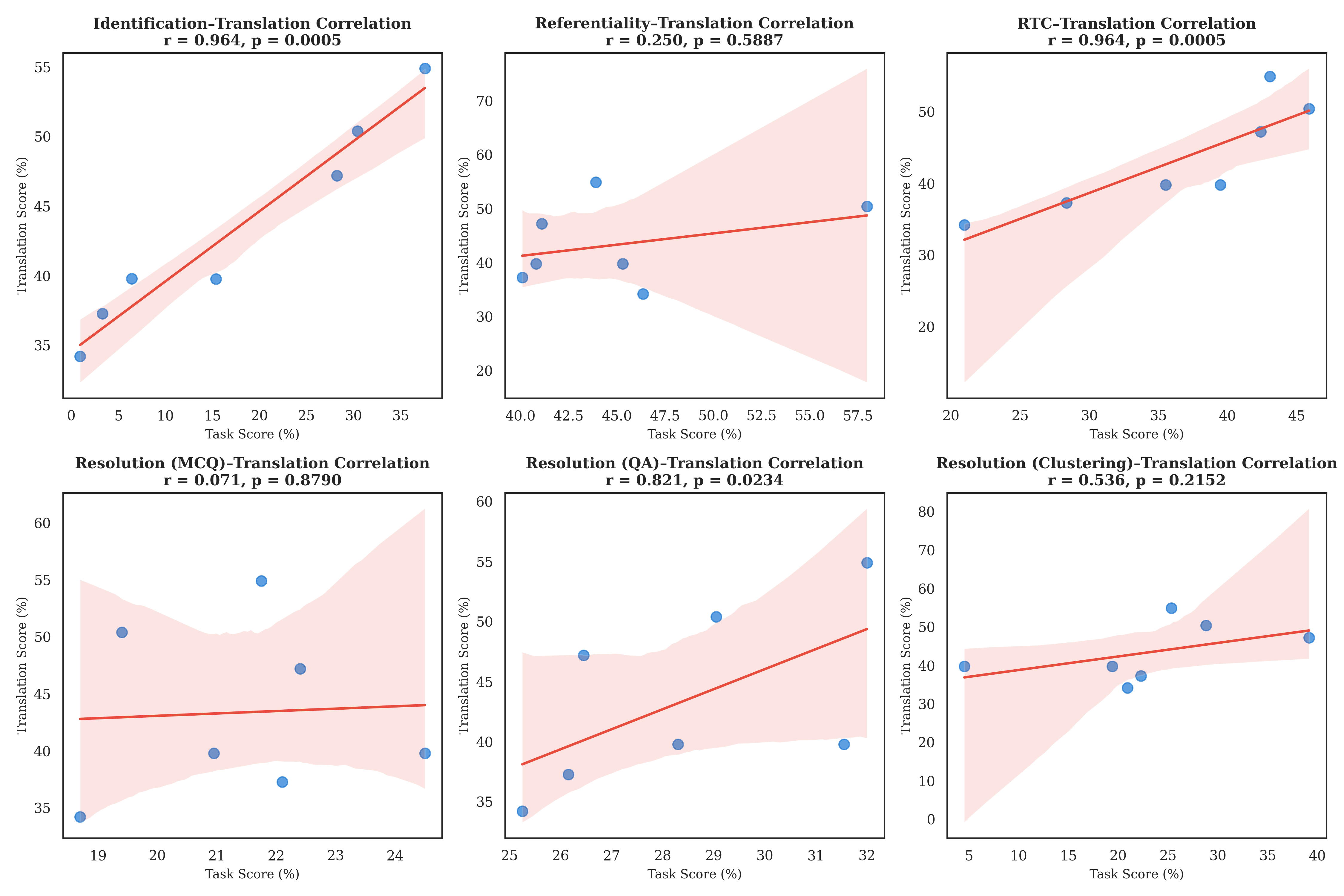}
    \caption{Relationships between ZP-aware translation performance and other task results, with Pearson correlation coefficients ($r$) and p-values ($p$) annotated in the title.}
    \label{fig:relevance}
\end{figure*}

Section~\ref{sec:result} shows that LLMs fail to correctly translate most ZPs in Chinese when given plain Chinese text as input. We therefore further investigate whether providing additional information about ZPs can improve performance. To this end, we consider two additional settings: ZP-aware translation, in which LLMs are still given plain Chinese text but are explicitly instructed that the text contains ZPs and that these should be handled carefully during translation; and translation with explicitly marked ZPs (referred to as `Oracle'), in which ZPs are directly annotated in the input.

Table~\ref{tab:translation} reports the results under these two settings. ZP-aware translation yields only marginal improvements, and only for sufficiently large models (e.g., DeepSeek-V3.2, GPT-4.1-mini) and reasoning-oriented models (e.g., DeepSeek-R1 and GPT-o4-mini); for smaller LLMs, it has negative effects. In contrast, the oracle setting, in which ZPs are explicitly marked, leads to substantial improvements for all models. Notably, reasoning-oriented LLMs achieve translation accuracies reaching 79\% under this setting. These findings suggest that explicitly providing the locations of ZPs, together with strong reasoning capability, substantially helps LLMs translate Chinese ZPs more accurately.

Table~\ref{tab:translation_example} presents example translations produced by DeepSeek-R1 under the three settings. When given plain text as input, the LLM can correctly handle almost no ZPs. Moreover, its difficulty in handling ZPs also leads to errors in translating some overt pronouns, for example, translating 自己 as `I' instead of `himself'. As more explicit information is provided, the model is able to translate a larger number of ZPs and do so with greater accuracy.

\paragraph{Relationships Between the Tasks.}

Our second research question concerns how LLMs’ performance on the downstream task (i.e., translation) relates to their performance on upstream tasks. Because LLMs perform uniformly poorly on ZP translation when given plain inputs (Table~\ref{tab:result}), we instead focus on ZP-aware translation and examine which capabilities enable LLMs to translate Chinese ZPs more effectively. Figure~\ref{fig:relevance} illustrates the relationships between ZP-aware translation and the other tasks considered in this study, along with the corresponding Pearson correlation coefficients for each task pair.

The results show that ZP identification is highly correlated with ZP-aware translation ($r = 0.964$). This finding is intuitive, as the oracle translation experiment already demonstrates that knowing the locations of ZPs can substantially improve translation quality. Referential type classification is also strongly correlated with ZP-aware translation. By contrast, referentiality classification exhibits only a weak correlation ($r = 0.250$), largely because nearly all LLMs perform poorly on this task, resulting in limited variance across models.

Among the three resolution tasks, QA-based resolution shows the strongest correlation with ZP-aware translation. This is reasonable as both QA-based resolution and translation require models to locate and extract antecedents. In contrast, ZP-aware translation shows little to no relationship with MCQ-based resolution, which, as discussed in Section~\ref{sec:result}, can be primarily attributed to LLMs’ poor performance on conversational data.

\section{Combining Referentiality and Referential Type Classification.} \label{sec:appendix_combine}
\input{tab/4way}
Given that referentiality can be viewed as a subcomponent of referential type classification, we investigate whether explicitly classifying referential types can help LLMs identify non-referential ZPs. To this end, we introduce a non-referential label in addition to the three original referential categories, resulting in a four-way classification variant of the referential type classification task.

The results of these additional comparative experiments, reported in Table~\ref{tab:4way}, reveal a clear bonus in the 4-way setting. Most LLMs show substantial improvements in non-referential F1, suggesting that introducing explicit categorical alternatives for referential ZPs provides a clearer contrastive boundary for filtering out structurally omitted arguments.

However, this improvement comes with a notable trade-off. Performance on the original three referential categories consistently declines, as reflected by the negative deltas in 3-way/4-way. This pattern indicates that adding the non-referential class complicates the decision boundaries among anaphora, cataphora, and deixis. The trade-off is particularly pronounced for stronger models such as DeepSeek-V3.2 and GPT-o4-mini, where non-referential accuracy even decreases in the 4-way setting, suggesting increased confusion among finer-grained referential distinctions.

\section{Influence of the Context Size} \label{sec:appendix_size}

\begin{figure*}
    \centering
    \includegraphics[scale=0.35]{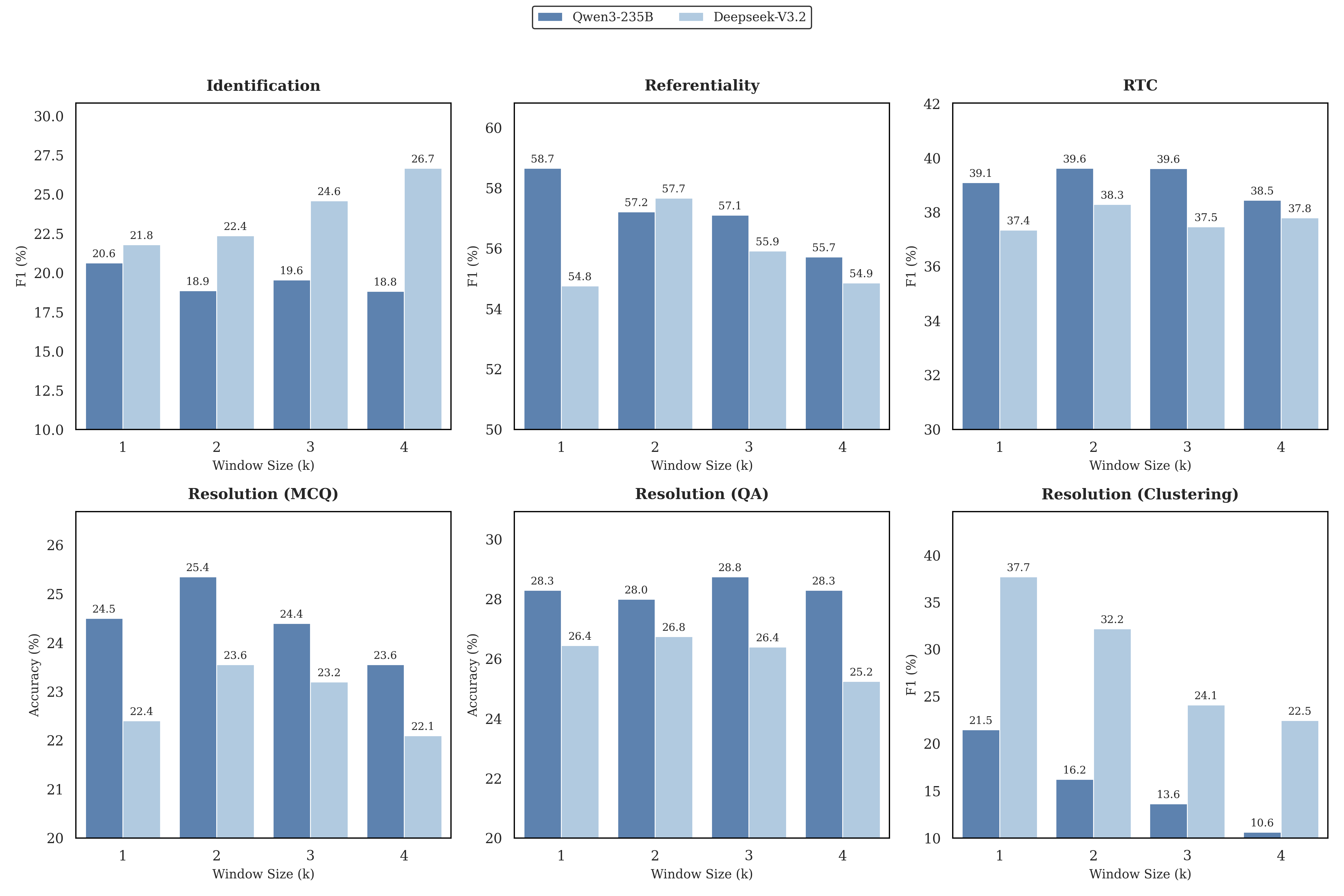}
    \caption{Performance comparison of Qwen3-235B and DeepSeek-V3.2 across different context window sizes ($k=1$ to $k=4$). Metrics are reported in F1-score (\%) for identification and classification tasks, and Accuracy (\%) for Resolution MCQ and QA.}
    \label{fig:window_size}
\end{figure*}

The understanding of ZP depends largely on discourse context, as the intended referent is often located outside the immediate sentence. In our main experiments, we provide the entire document within the prompt for all tasks except for resolution and translation. However, we are curious about whether the specific volume of the surrounding context impacts LLM performance. To investigate this, we evaluate the models across varying context window sizes ($k$), ranging from 1 to 4. This $k$ value represents the number of sentences provided both before and after the target sentence containing the ZP. We repeated this analysis across four core tasks: Identification, Referentiality, Referential Type Classification (RTC), and Resolution. Figure~\ref{fig:window_size} records the results for Qwen-235B and DeepSeek-V3.2.


For Qwen3-235B, which exhibits relatively weak capability in handling Chinese zero pronouns, increasing the context size, and thus providing more discourse information, has little effect on most tasks. For referentiality classification, MCQ-based resolution, and clustering-based resolution, larger context sizes even lead to performance degradation. These results suggest that, when a model has limited ability to process long discourse and reason over it, additional contextual information may instead introduce confusion rather than provide useful cues.
 
For DeepSeek-V3.2, performance improves significantly on the ZP identification task as the context window widens. In contrast, larger contexts have a negative effect on clustering-based resolution, which is expected: broader contexts introduce more mentions that must be clustered, thereby increasing task difficulty. For the remaining tasks, context size does not have a significant impact on DeepSeek-V3.2’s performance.

\section{Conclusion}

This paper presents a systematic study of Large Language Models’ (LLM) ability to process Chinese Zero Pronouns (ZPs) through a sequence of linguistically motivated tasks. We decompose ZP understanding into five stages—identification, referentiality classification, referential type classification, resolution, and translation—and evaluate a diverse set of LLMs across all stages. This design enables a fine-grained analysis of ZP processing as a discourse-level phenomenon and allows us to examine how model size and explicit reasoning capabilities influence performance at different points in the processing pipeline.

Our results show that Chinese ZPs remain highly challenging for current LLMs. Across all tasks, performance is consistently low, with particularly severe difficulties observed in upstream tasks such as identification and referentiality classification. Although larger and reasoning-oriented models achieve better results, these improvements are limited and do not close the gap. Further analyses show that, when explicit ZP information is provided, models with stronger performance on upstream tasks, particularly identification and referential type classification, exhibit larger gains in translation accuracy. This highlights the close relationship between upstream ZP processing and downstream translation performance. We hope that this study can serve as a benchmark for evaluating discourse-level reasoning capabilities in Chinese and for guiding future research in this direction.

\section*{Limitations}

On the one hand, although we evaluate a diverse set of large language models with varying sizes and reasoning capabilities, we do not include the most recent state-of-the-art reasoning models, such as GPT-o5. This decision is primarily driven by practical cost considerations: even GPT-o4-mini is already approximately ten times more expensive than DeepSeek-R1 in our experimental setting. As a result, our findings may not fully reflect the upper bound of performance achievable by the most advanced and costly models.

On the other hand, our investigation is limited to Chinese. While Chinese provides a particularly challenging and well-studied test case for zero pronouns, it represents only one instance of a broader class of pro-drop languages. Other languages, such as Arabic, Japanese, and Korean, exhibit different syntactic and discourse properties that may interact with zero pronoun processing in distinct ways. Extending the current framework to these languages is an important direction for future work and would help assess the generality of our findings across pro-drop phenomena. In addition, the present study focuses primarily on the comprehension and resolution of zero pronouns, rather than on pro-drop in language production. Future work could further examine whether and under what conditions LLMs appropriately omit overt pronouns when generating text in pro-drop languages~\citep{chen-etal-2018-modelling,chen2022computational}, and whether their production patterns align with native-speaker norms across different discourse contexts.

\paragraph{Use of AI Assistants.} In this work, we used GPT-5.2 to assist with refining the language of the paper and to help write code for data preprocessing, result analysis, and figure drawing. All experimental design decisions, analyses, and interpretations were made by the authors.

\section*{Acknowledgment}

This work was supported by the MOE (Ministry of Education in China) Project of Humanities and Social Sciences (Project No.25YJC740005), the National Language and Character Research Base  (Project No.ZDI145-168), and Fundamental Research Funds for the Central Universities, Academy of Frontier Interdisciplinary Research, Central China Normal University (Project No.JC2026PT-004).

\bibliography{custom,anthology-1,anthology-2}

\appendix

\section{Data Preprocessing} \label{sec:app_process}

This appendix details how we extract labels from the OntoNotes dataset for the five tasks considered in this study. In OntoNotes, Zero Pronouns (ZPs) are marked using the token `*pro*', and annotations of coreference chains and referentiality are stored in the .onf and .coref files. our analysis relies on *pro* annotations in OntoNotes, which operationalise omitted pronominal arguments in the treebank. Because pure discourse-level topic ellipsis that does not involve a missing argument is generally not annotated in this scheme, such cases are not included in our evaluation, effectively distinguishing ZP–like omissions from topic-only ellipsis.

\input{tab/data_statistics}

\paragraph{ZP Identification.} For the first task, we focus on identifying the positions of all ZPs, which are explicitly annotated as `*pro*' in the .coref files. During preprocessing, we remove word segmentation information and other non-informative annotations while preserving the `*pro*' tokens, which serve as gold labels for ZP identification.

\paragraph{Referentiality Classification.} This task aims to determine whether a given zero pronoun is referential in its context. We derive referentiality labels from the coreference annotations: a ZP is labelled as \textit{referential} if it participates in a valid coreference chain (annotated with \textit{COREF} in the .coref files), and as \textit{non-referential} otherwise.

\paragraph{Referential Type Classification.} For referential ZPs identified in the preceding task, we further classify their referential type in discourse as \textit{anaphora}, \textit{cataphora}, or \textit{deixis}. Deictic ZPs are identified by inspecting their corresponding coreference chains in the .onf files using the \textit{COREF ID} labels from the .coref files: chains composed exclusively of *pro* tokens and deictic expressions (e.g., first-, second-, or third-person pronouns, or demonstratives) are labelled as deixis. For the remaining cases, the referential type is determined by the linear order of mentions within the coreference chain: if an explicit antecedent precedes the target ZP, it is labelled as anaphora; otherwise, it is labelled as cataphora.

\paragraph{Resolution.} The goal of resolution is to link a referential ZP to its antecedent entity in the discourse. Data preprocessing for this task involves constructing candidate sets and gold-standard labels. For each target ZP, candidate noun phrases are extracted from noun-related nodes in the syntactic parse trees provided in the .onf files, and gold antecedents are obtained from the target ZP’s coreference chain, restricted to explicit entities within the context. To accommodate our three evaluation settings, we apply task-specific data transformations:
(1) For the \textbf{MCQ-based method}, the full list of noun-phrase candidates extracted from the parse trees is provided, and the gold labels are defined as the subset of these candidates that belong to the reference chain.
(2) For the \textbf{QA-based method}, no candidate list is given; the gold labels are identical to those used in the MCQ-based setting.
(3) For the \textbf{Clustering-based method}, while the candidate mentions and gold antecedents remain the same as in the previous settings, the gold labels are represented as a gold cluster of character-level pairs. Each pair consists of a character from a gold antecedent mention and its corresponding absolute character-level index in the context.

\paragraph{Translation.} For this task, we preserve only the plain text, without any word segmentation, to construct both simple and ZP-aware prompt variants. We retain ZP markers when constructing the oracle version of the prompts. In addition, we preserve the \textit{COREF ID} of each ZP and phrase from the .coref files to facilitate comparison during human evaluation.

\section{Data Statistics} \label{sec:app_statistics}

We summarise here the statistics of the sampled data derived from the Chinese portion of the OntoNotes 5.0 dataset. The final dataset contains 3,607 ZPs spanning six genres. Table~\ref{tab:data_statistics} reports the distribution of instances for each task across these genres. Notably, data density varies substantially by genre: spoken-domain categories such as BC and TC exhibit a much higher frequency of ZPs than more formal domains such as NW.

\input{tab/appendix_pr}

\section{Implementation Details} 

There are nine prompt templates elaborately designed for all methods in five tasks (see Table~\ref{tab:prompt_templates} for the full templates). Before prompting LLMs to execution, prompt construction based on these templates mainly differs in how contextual information and supervision signals are presented according to the objective of the task. For identification, referentiality classification, referential type classification, QA-based resolution, and translation, prompts are constructed by embedding the plain or slightly processed discourse directly into fixed templates. 

In contrast, MCQ-based and clustering-based resolution necessitate more complex construction beyond simple contextual insertion due to the requirement for antecedent selection. Both utilise candidate noun phrases extracted via syntactic tree parsing. However, they differ in presentation: the former provides a discrete list of candidate noun phrases as indexed options, while the latter employs a '<>‘ marker to disclose each candidate noun phrase within the discourse.  

\section{Complementary Results}

This appendix provides complementary results of our experiments, including precision and recall scores, per-genre results, and per-class results.

\subsection{Precision and Recall Scores} \label{sec:appendix_pr}

While the main text focuses on the F1-score as a harmonic mean of performance, the trade-off between precision and recall reveals the behavioural tendencies of different LLMs. Detailed precision and recall results in identification, referentiality classification, referential type classification and clustering-based resolution are provided in Table~\ref{tab:apprendix_pr}.

\subsection{Per-genre Results} \label{sec:appendix_genre}

\input{tab/appendix_genre}

In our experiments, we also record performance metrics broken down by genre. As shown in Table~\ref{tab:appendix_genre}, LLM performance varies substantially across the six genres.

\subsection{Per-class Results} \label{sec:appendix_class}
\input{tab/appendix_classes}
To further expose the performance disparities and inherent biases that are otherwise masked by aggregate F1-scores, detailed results on each class are shown in Table~\ref{tab:appendix_classes}.

\paragraph{Referentiality Classification.} We report the F1-score for both positive (referential) and negative (non-referential) classes in this task, finding that while models maintain a relatively high and stable F1-score for the positive class (ranging from 70.52 to 72.42), performance on the negative class is generally low, except GPT-o4-mini as an outlier achieving a significantly higher F1 score non-referential of 44.84 (others ranging from 0.14 to 13.20).

\paragraph{Referential type classification.} The class-level breakdown for this task reveals that the position of the antecedent makes a tremendous difference. Anaphora remains the strongest category across all models, likely due to the explicit presence of antecedents in the preceding context. In contrast, cataphora and deixis show significantly lower scores, reflecting the difficulty models face when antecedents appear in the following context or must be inferred from the situational environment.

\section{A Case Study of Reasoning-oriented LLMs} \label{sec:appendix_reasoning_example}

In our main studies, Table~\ref{sec:result} reveals that DeepSeek-R1 substantially outperforms all other models on the ZP identification task, achieving the highest F1-score by a clear margin. To better understand the sources of these gains, we conduct targeted case studies on these two models, focusing on how their explicit reasoning traces and implicit inference patterns contribute to task-specific advantages.

\input{tab/reasoning_example}

To analyse why DeepSeek-R1 excels at identifying zero pronouns, we examine its reasoning behaviour on representative examples. As shown in Table~\ref{tab:reasoning_example}, the model adopts a structured reasoning process that consists of (1) clause-level segmentation to isolate predicates, and (2) a valency-based audit that checks whether each predicate’s required arguments are overtly realised. This predicate-centric strategy allows the model to accurately detect structurally licensed omissions while avoiding overgeneration in adverbial or discourse-only fragments.

\section{Error Analysis} \label{sec:appendix_error}

Table~\ref{tab:error_analysis} shows analyses of two LLMs (Qwen3-8B and DeepSeek-V3.2) on Identification, Referentiality Classification, Referential Type Classification, and ZP Resolution.

\onecolumn
\input{tab/error_analysis}
\clearpage
\twocolumn

\input{tab/prompt_templates}

\end{CJK}
\end{document}

%% file: tab/result.tex
\begin{table*}[t]
    \centering
    \small
    \begin{tabular}{lcccccccc}
        \toprule
& \textbf{Ident.} & \textbf{Ref.} & \textbf{RTC} & \textbf{MCQ} & \textbf{QA} & \textbf{Clustering} & \textbf{Translation} & \textbf{Borda} \\
Model & F1 & F1 & F1 & Accuracy & Accuracy & F1 & Acc. & Count \\
\midrule
Qwen3-4B & 3.19 & 40.01 & 20.96 & 18.70 & 25.25 & 22.63 & 35.32$^{A,B}$ & 6 \\
Qwen3-8B & 3.32 & 35.64 & 28.36 & 22.10 & 26.15 & 24.67 & 37.50$^{A,B}$ & 13 \\
Qwen3-235B & 15.38 & 37.98 & 35.52 & \textbf{24.50} & 28.30 & 21.49 & \underline{39.78}$^{A,B}$ & 21 \\
DeepSeek-V3.2 & 28.24 & 38.34 & 42.40 & \underline{22.40} & 26.45 & \textbf{37.70} & \textbf{43.59}$^{A}\;\;\;$ & \underline{29} \\
DeepSeek-R1 & \textbf{37.58} & 39.87 & \underline{43.07} & 21.75 & \textbf{32.00} & 23.65 & 35.56$^{B\;\;\;}$ & 27 \\
GPT-4.1-mini & 6.43 & \underline{41.86} & 39.47 & 20.95 & \underline{31.55} & 4.68 & 38.46$^{A,B}$ & 21 \\
GPT-o4-mini & \underline{30.41} & \textbf{58.63} & \textbf{45.89} & 19.40 & 29.05 & \underline{26.46} & 38.10$^{A,B}$ & \textbf{30} \\
        \bottomrule
    \end{tabular}
    \caption{Main results of our study, in which `Ident.' means identification, `Ref.' means Referentiality Classification, `RTC' means Referential Type Classification. Best results are \textbf{boldfaced}, whereas the second best are \underline{underlined}. Per column, results that have no superscript letters in common are significantly different from each other (p < 0.05). Statistical Testing was done using the McNemar test.}
    \label{tab:result}
\end{table*}

%% file: tab/translation.tex
\begin{table}[t]
    \centering
    \small
    \begin{tabular}{lll}
        \toprule
Model & ZP-aware & Oracle \\
\midrule
Qwen3-4B & 34.20 $_{\color{red} (-1.12)\;\;}^{D}$ & 43.20 $_{\color{Green} (+7.88)}\;^{D}$ \\
Qwen3-8B & 37.27 $_{\color{Red} (-0.23)\;\;}^{D}$ & 54.20 $_{\color{Green} (+16.70)}^{C,D}$ \\
Qwen3-235B & 39.77 $_{\color{red} (-0.01)\;\;}^{C,D}$ & 59.44 $_{\color{Green} (+19.66)}^{C}$ \\
DeepSeek-V3.2 & 47.19 $_{\color{Green} (+3.60)\;\;}^{B,C}$ & 65.43 $_{\color{Green} (+21.84)}^{B}$ \\
DeepSeek-R1 & 54.89 $_{\color{Green} (+19.33)}^{A}$ & 79.01 $_{\color{Green} (+43.45)}^{A}$ \\
GPT-4.1-mini & 39.78 $_{\color{Green} (+1.32)\;\;}^{C,D}$ & 63.00 $_{\color{Green} (+24.54)}^{B}$ \\
GPT-o4-mini & 50.40 $_{\color{Green} (+12.30)}^{A,B}$ & 77.83 $_{\color{Green} (+39.73)}^{A,B}$ \\
        \bottomrule
    \end{tabular}
    \caption{Zero-pronoun translation accuracies under the ZP-aware and Oracle settings. The number following each score indicates the change relative to the ZP translation accuracy obtained with plain inputs, as reported in Table~\ref{tab:result}.}
    \label{tab:translation}
\end{table}

%% file: tab/translation_example.tex
\begin{table}[t]
    \centering
    \small
    \begin{tabular}{lp{5.3cm}}
        \toprule
Input & <pro>讲啊.<pro>会说<pro>想想舅舅了。现在反正自己挺爱说的了。\\
& 呃他--他是不是都忘了我,忘了我长啥样,肯定的哈? \\
& <pro>没忘,<pro>一来电话咱们老拿照片看。\\
\midrule
Plain & \textcolor{red}{Say}, {\color{red}I} was just thinking of Uncle. {\color{red}I}'ve become quite talkative lately anyway. \\
& Uh, has he—has he forgotten about me, forgotten what I look like? For sure, right? \\
& No, {\color{Green}he} hasn't. Whenever {\color{red}he} calls, we always look at photos. \\ 
ZP-aware & \textcolor{red}{I} say, {\color{Green}he} will say {\color{Green}he} misses uncle. Now anyway, he himself quite loves to talk. \\
& Uh he--has he forgotten me, forgotten what I look like, for sure huh? \\
& Not forgotten, \textcolor{red}{whenever there's a call} we always take out the photos to look.\\
Oracle & \textcolor{Green}{He} is talking. \textcolor{Green}{He} says \textcolor{Green}{he} misses uncle. Now anyway, he loves to talk. \\
& Uh, he—has he forgotten me, forgotten what I look like, for sure? \\
& \textcolor{Green}{He} hasn't forgotten, as soon as \textcolor{red}{he} calls, we always take out the photos to look. \\
        \bottomrule
    \end{tabular}
    \caption{Example translations produced by DeepSeek-R1 under the three translation settings. Correctly translated zero pronouns are highlighted in green, while incorrect translations resulting from misinterpretation of zero pronouns are highlighted in red.}
    \label{tab:translation_example}
\end{table}

%% file: tab/4way.tex
\begin{table*}[t]
    \centering
    \small
    \begin{tabular}{llllll}
        \toprule
Model & 3-way & 4-way & 3-way/4-way & Non-Referential & Non-referential/4-way \\
\midrule
Qwen3-4B & 20.96 & 10.90 & 8.85 $_{\color{red} (-12.11)}$ & 9.46 & 17.04 $_{\color{Green} (+7.57)}$\\
Qwen3-8B & 28.36 & 19.76 & 20.00 $_{\color{red} (-8.36)}$ & 0.14 & 19.03 $_{\color{Green} (+18.90)}$\\
Qwen3-235B & 35.52 & 21.37 & 25.57 $_{\color{red} (-9.95)}$ & 4.43 & 8.78 $_{\color{Green} (+4.35)}$\\
DeepSeek-V3.2 & 42.40 & 15.10 & 20.01 $_{\color{red} (-22.39)}$ & 5.68 & 0.37 $_{\color{red} (-5.30)}$\\
DeepSeek-R1 & \underline{43.07} & \underline{28.37} & \textbf{34.44} $_{\color{red} (-8.63)}$ & 8.04 & 10.17 $_{\color{Green} (+2.13)}$\\
GPT-4.1-mini & 39.47 & 25.10 & 25.17 $_{\color{red} (-14.30)}$ & \underline{13.20} & \underline{24.91} $_{\color{Green} (+11.71)}$\\
GPT-o4-mini & \textbf{45.89} & \textbf{32.22} & \underline{31.01} $_{\color{red} (-14.88)}$ & \textbf{44.82} & \textbf{35.87} $_{\color{red} (-8.98)}$\\
        \bottomrule
    \end{tabular}
    \caption{Comparative analysis combining referentiality and referential type classification. 3-way and 4-way denote the F1 scores of referential type classification without and with non-referential as an additional category, respectively. Non-referential refers to performance on the non-referential class in the referentiality classification task. 3-way/4-way indicates the averaged F1 score over the original three referential types within the 4-way classification setting, while Non-referential/4-way denotes the F1 score for the non-referential class in the 4-way setting. Subscripted numbers indicate changes relative to the original task.}
    \label{tab:4way}
\end{table*}

%% file: tab/data_statistics.tex
\begin{table*}[t]
    \centering
    \small
    \begin{tabular}{llccccccc}
        \toprule
        & & \textbf{bc} & \textbf{bn} & \textbf{mz} & \textbf{nw} & \textbf{tc} & \textbf{wb} & \textbf{total} \\
        \midrule
        \multicolumn{2}{l}{\textbf{Identification}} & 1158 & 763 & 201 & 94 & 724 & 667 & 3607 \\
        \midrule
        \multirow{2}{*}{\textbf{Referentiality}} & Referential & 602 & 435 & 179 & 87 & 339 & 358 & 2000 \\
        & Non-referential & 556 & 328 & 22 & 7 & 385 & 309 & 1607 \\
        \midrule
        \multirow{4}{*}{\textbf{Referential Type}} & Anaphora & 403 & 360 & 140 & 84 & 103 & 168 & 1258 \\
        & Cataphora & 25 & 18 & 11 & 3 & 12 & 13 & 82 \\
        & Deixis & 174 & 33 & 10 & 0 & 224 & 48 & 489 \\
        \midrule
        \multirow{3}{*}{\textbf{Resolution}} & MCQ & 602 & 435 & 179 & 87 & 339 & 358 & 2000 \\
        & QA & 602 & 435 & 179 & 87 & 339 & 358 & 2000 \\
        & Clustering & 1397 & 3436 & 757 & 367 & 377 & 1365 & 7699 \\
        \midrule
        \multicolumn{2}{l}{\textbf{Translation}} & 39 & 45 & 48 & 31 & 49 & 57 & 269 \\
        \bottomrule
    \end{tabular}
    \caption{Detailed statistics of ZP in the dataset per task across six genres.}
    \label{tab:data_statistics}
\end{table*}

%% file: tab/appendix_pr.tex
\begin{table*}[t]
    \centering
    \small 
    \begin{tabular}{lcccccccc}
        \toprule
& \multicolumn{2}{c}{\textbf{Identification}} & \multicolumn{2}{c}{\textbf{Referentiality}} & \multicolumn{2}{c}{\textbf{Referential Type}} & \multicolumn{2}{c}{\textbf{Clustering}} \\\cmidrule(lr){2-3} \cmidrule(lr){4-5} \cmidrule(lr){6-7} \cmidrule(lr){8-9}
Model & P & R & P & R & P & R & P & R \\
\midrule
Qwen3-4B & 9.41 & 1.92 & 53.91 & 50.67 & 37.23 & 35.43 & 27.91 & 19.03 \\
Qwen3-8B & 6.00 & 2.29 & 34.80 & 49.87 & 32.04 & 31.78 & 29.90 & \underline{21.00} \\
Qwen3-235B & 11.96 & 21.56 & 61.65 & 50.70 & 35.41 & 35.63 & 50.14 & 13.68 \\
DeepSeek-V3.2 & 25.77 & \underline{31.24} & 58.13 & 50.71 & 43.79 & 42.72 & 52.77 & \textbf{29.33} \\
DeepSeek-R1 & \underline{33.10} & \textbf{43.47} & \textbf{66.02} & \underline{51.58} & \textbf{48.69} & \underline{42.75} & \textbf{67.21} & 14.35 \\
GPT-4.1-mini & 13.32 & 4.24 & 54.54 & 51.09 & 40.96 & 40.65 & 40.64 & 2.48 \\
GPT-o4-mini & \textbf{38.84} & 25.00 & \underline{64.43} & \textbf{60.18} & \underline{45.88} & \textbf{46.66} & \underline{67.07} & 16.48 \\
        \bottomrule
    \end{tabular}
    \caption{Precision and recall scores in identification, referentiality classification, referential type classification and clustering-based resolution.}
    \label{tab:apprendix_pr}
\end{table*}

%% file: tab/appendix_genre.tex
\begin{table*}[p]
    \small
    \centering
    \begin{tabular}{lccccccc}
        \toprule
& \textbf{Ident.} & \textbf{Ref.} & \textbf{RTC} & \textbf{MCQ} & \textbf{QA} & \textbf{Clustering} & \textbf{Translation} \\
Model & F1 & F1 & F1 & Acc. & Acc. & F1 & Acc. \\
\midrule
\multicolumn{8}{c}{\textbf{bc (Broadcast Conversation)}} \\ \midrule
Qwen3-4B & 0.00 & 34.37 & 18.74 & 10.13 & 23.42 & 19.19 & 28.21 \\
Qwen3-8B & 0.00 & 34.21 & 33.23 & 10.63 & 23.59 & 20.39 & 28.21 \\
Qwen3-235B & 16.16 & 38.95 & 37.05 & \textbf{12.13} & 26.08 & 21.07 & 33.33 \\
DeepSeek-V3.2 & 27.02 & 36.14 & \underline{38.92} & \underline{11.46} & 23.42 & \underline{36.43} & \textbf{41.03} \\
DeepSeek-R1 & \textbf{45.33} & 37.52 & 37.48 & \underline{11.46} & \underline{30.57} & \textbf{38.72} & 33.33 \\
GPT-4.1-mini & 2.94 & \underline{40.67} & 34.14 & 9.14 & \textbf{32.89} & 8.47 & \underline{38.46} \\
GPT-o4-mini & \underline{34.67} & \textbf{57.25} & \textbf{43.40} & 8.47 & 29.07 & 35.92 & 33.33 \\
        
\midrule
\multicolumn{8}{c}{\textbf{bn (Broadcast News)}}  \\\midrule
Qwen3-4B & 2.74 & 37.70 & 16.50 & 32.64 & 26.90 & 25.02 & 26.67 \\
Qwen3-8B & 2.58 & 35.82 & 26.20 & \underline{37.93} & 30.81 & \underline{27.38} & 33.33 \\
Qwen3-235B & 9.46 & 36.61 & 39.75 & \textbf{42.99} & 31.72 & 22.24 & \textbf{44.44} \\
DeepSeek-V3.2 & \underline{18.19} & 37.86 & 34.99 & 37.01 & \underline{32.18} & \textbf{37.04} & \underline{35.56} \\
DeepSeek-R1 & \textbf{18.57} & \underline{42.87} & 38.71 & 33.79 & \textbf{33.33} & 14.55 & 23.81 \\
GPT-4.1-mini & 5.63 & 40.00 & \underline{43.03} & 37.24 & 29.66 & 3.76 & 31.11 \\
GPT-o4-mini & 15.04 & \textbf{62.18} & \textbf{43.99} & 32.18 & 26.67 & 20.43 & 33.33 \\

\midrule
\multicolumn{8}{c}{\textbf{mz (Magazine Articles)}} \\\midrule
Qwen3-4B & 0.00 & 46.45 & 6.33 & 34.08 & 25.70 & 18.11 & \underline{50.00} \\
Qwen3-8B & 0.00 & 46.45 & 15.01 & \underline{44.13} & 21.79 & \underline{19.86} & 49.02 \\
Qwen3-235B & 10.23 & 0.00 & 35.09 & \textbf{45.25} & 30.73 & 18.20 & 39.58 \\
DeepSeek-V3.2 & 24.84 & 46.86 & 34.12 & 41.90 & \underline{32.96} & \textbf{33.59} & \textbf{53.85} \\
DeepSeek-R1 & \textbf{34.48} & \underline{54.23} & \textbf{56.66} & 41.34 & \textbf{33.52} & 10.94 & 44.23 \\
GPT-4.1-mini & 5.83 & 46.18 & 33.24 & 39.11 & 30.17 & 4.39 & 40.38 \\
GPT-o4-mini & \underline{33.33} & \textbf{54.44} & \underline{45.25} & 40.22 & 27.93 & 13.78 & 36.54 \\

\midrule
\multicolumn{8}{c}{\textbf{nw (Newswire)}} \\\midrule
Qwen3-4B & 7.46 & \textbf{55.49} & 7.64 & 31.03 & 19.54 & 19.36 & 25.81 \\
Qwen3-8B & 4.85 & 47.78 & 24.02 & 36.78 & 22.99 & 21.05 & 22.58 \\
Qwen3-235B & 5.53 & 0.00 & 31.30 & 43.68 & \textbf{34.48} & 14.01 & 25.81 \\
DeepSeek-V3.2 & 12.62 & \underline{48.86} & 32.70 & 41.38 & 26.44 & \textbf{43.02} & \textbf{38.71} \\
DeepSeek-R1 & \underline{18.99} & 48.24 & \textbf{43.15} & \underline{45.98} & \underline{31.03} & 12.94 & 19.35 \\
GPT-4.1-mini & 10.19 & 47.49 & 27.74 & \textbf{47.13} & 25.29 & 2.10 & 25.81 \\
GPT-o4-mini & \textbf{20.00} & 48.44 & \underline{40.13} & 40.23 & 24.14 & \underline{25.78} & \underline{29.03} \\

\midrule
\multicolumn{8}{c}{\textbf{tc (Telephone Conversation)}} \\\midrule
Qwen3-4B & 3.97 & 31.89 & 26.52 & \textbf{5.90} & 29.20 & 15.18 & 38.78 \\
Qwen3-8B & 8.61 & 31.89 & 27.95 & \underline{5.61} & 30.68 & 13.52 & 40.82 \\
Qwen3-235B & 26.14 & 32.05 & \underline{34.39} & \textbf{5.90} & 25.96 & 14.44 & \underline{42.86} \\
DeepSeek-V3.2 & 39.2 & 34.92 & 31.61 & \textbf{5.90} & 22.12 & \underline{41.03} & 40.82 \\
DeepSeek-R1 & \textbf{56.85} & 32.97 & 33.59 & 5.61 & \underline{33.04} & 34.91 & 36.73 \\
GPT-4.1-mini & 11.63 & \underline{39.44} & 31.24 & 4.72 & \textbf{37.17} & 4.57 & 40.82 \\
GPT-o4-mini & \underline{51.35} & \textbf{56.80} & \textbf{37.35} & 4.43 & 31.56 & \textbf{41.26} & \textbf{44.90} \\

\midrule
\multicolumn{8}{c}{\textbf{wb (Web Blogs)}} \\\midrule
Qwen3-4B & 0.00 & \textbf{58.04} & 17.65 & 17.60 & 23.74 & 26.18 & 36.84 \\
Qwen3-8B & 3.01 & 37.46 & 28.49 & 23.18 & 23.46 & 29.47 & 42.11 \\
Qwen3-235B & 14.80 & 38.30 & 29.58 & \textbf{25.42} & 27.37 & 25.77 & \underline{45.61} \\
DeepSeek-V3.2 & 36.58 & 38.41 & 33.74 & \underline{24.30} & 25.42 & \textbf{40.53} & \textbf{47.37} \\
DeepSeek-R1 & \underline{62.86} & 43.36 & \textbf{50.54} & 24.02 & \textbf{32.40} & 31.08 & \underline{45.61} \\
GPT-4.1-mini & 7.24 & 42.50 & \underline{41.94} & 20.95 & 28.49 & 3.96 & \textbf{47.37} \\
GPT-o4-mini & \textbf{66.67} & \underline{56.02} & 39.36 & 20.95 & \underline{31.29} & \underline{31.14} & \underline{45.61} \\

        \bottomrule
    \end{tabular}
    \caption{Detailed performance across all six genres.}
    \label{tab:appendix_genre}
\end{table*}

%% file: tab/appendix_classes.tex
\begin{table*}[]
    \centering
    \small
    \begin{tabular}{lccccc}
        \toprule
        & \multicolumn{2}{c}{\textbf{Referentiality}} & \multicolumn{3}{c}{\textbf{Referential Type Classification}} \\ \cmidrule(lr){2-3} \cmidrule(lr){4-6}
        Model & Referential & Non-referential & Anaphora & Cataphora & Deixis \\
        \midrule
        Qwen3-4B & 70.56 & 9.46 & 15.43 & 2.75 & \underline{44.71} \\
        Qwen3-8B & 71.15 & 0.14 & 67.55 & 5.06 & 12.47\\
        Qwen3-235B & 71.53 & 4.43 & \textbf{86.43} & 3.35 & 16.77 \\
        DeepSeek-V3.2 & 71.01 & 5.68 & \underline{81.07} & 0.00 & \textbf{46.13} \\
        DeepSeek-R1 & \underline{71.70} & 8.04 & 79.68 & \textbf{18.69} & 30.83 \\
        GPT-4.1-mini & 70.52 & \underline{13.20} & 75.10 & 9.49 & 33.82 \\
        GPT-o4-mini & \textbf{72.42} & \textbf{44.84} & 77.34 & \underline{17.56} & 42.78 \\
        \bottomrule
    \end{tabular}
    \caption{Class-level metrics of our study.}
    \label{tab:appendix_classes}
\end{table*}

%% file: tab/reasoning_example.tex
\begin{table*}[t]
    \small
    \centering
    \begin{tabular}{lp{12cm}} 
        \toprule
        \textbf{Input} & “试图近距离去开始于两年前的二零零三年，但是在迪斯尼的任何标志都还没有进入我们的视线时，车子就在去往迪斯尼的岔路口被保安人员拦了下来，” \\
        \midrule
        \textbf{Reasoning} & 任务是标注零代词缺失位置 ... \textbf{分句分析}：分句1“试图近距离去开始于...”结构不完整；分句2“在迪斯尼的任何标志都还没有进入我们的视线时”为状语从句；分句3“车子就在...被保安人员拦了下来”为主句 ... \textbf{判断省略}：分句1谓语动词“试图”需要主语但缺失，应插入\texttt{<pro>}；分句2主语为“迪斯尼的任何标志”，成分完整；分句3主语为“车子”，被动句结构完整，无省略 ... \textbf{最终判定}：参考“明天去爬山吗？”等主语省略示例，仅在分句1句首插入\texttt{<pro>} ... 保持原样输出。 \\
        \midrule
        \textbf{Output} & “\textbf{<pro>}试图近距离去开始于两年前的二零零三年，但是在迪斯尼的任何标志都还没有进入我们的视线时，车子就在去往迪斯尼的岔路口被保安人员拦了下来，” \\
        \bottomrule
    \end{tabular}
    \caption{A distilled reasoning trace from DeepSeek-R1 demonstrating syntactic valency auditing for ZP Identification. The model systematically identifies missing arguments while correctly rejecting complete clauses to avoid over-identification.}
    \label{tab:reasoning_example}
\end{table*}

%% file: tab/error_analysis.tex
{\small \begin{longtable}{lp{14cm}}
    \toprule
    \textbf{Task} & \textbf{Details} \\
    \midrule
    \endfirsthead
    \endhead
    \endfoot
    \textbf{Task 1} & \textbf{Identification} \\
    \midrule
    Example 1 & 前几天，又一位台湾明星离开。我之前没有听说过她，<pro>听说<pro>是个有才华的美女。<pro>非常惋惜。\\
    Qwen3-8b & 前几天，又一位台湾明星离开。<pro>我之前没有听说过她，听说是个有才华的美女。非常惋惜。\\
    & \textbf{Error Analysis}: Underdetermined. Tends to focus only on explicit subject transitions; lacks discourse-level subject recovery. \\
    DeepSeek & 前几天，又一位台湾明星离开。我之前没有听说过她，<pro>听说是个有才华的美女。<pro>非常惋惜。 \\
    & \textbf{Error Analysis}: Over-determined. Overly sensitive to pre-verbal positions, misjudging non-essential positions as ZPs. \\
    \midrule
    Example 2 & 所以，车子的侧面气囊和气帘全部打开了。这是<pro>第一次撞击护拦的时候打开的，第一次的撞击属于<pro>蹭护拦而已，人员本身受到的撞击力都非常小，而且<pro>有气帘的保护，所以肯定不会有大问题。\\
    Qwen3-8b & 所以，车子的侧面气囊和气帘全部打开了。这是第一次撞击护拦的时候打开的，第一次的撞击属于蹭护拦而已，人员本身受到的撞击力都非常小，而且有气帘的保护，所以肯定不会有大问题。\\
    & \textbf{Error Analysis}: Under-determined. Struggles to capture cross-clause shared subject relations in long, complex event descriptions. \\
    DeepSeek & 所以，车子的侧面气囊和气帘全部打开了。这是第一次撞击护拦的时候打开的，第一次的撞击属于蹭护拦而已，人员本身受到的撞击力都非常小，而且<pro>有气帘的保护，所以<pro>肯定不会有大问题。。\\
    & \textbf{Error Analysis}: Mixed Error. Over-completes inferential semantic positions due to semantic inertia. \\
    
    \midrule
    \textbf{Task 2} & \textbf{Referentiality Classification} \\
    \midrule
    Example 1 & 中国全国人民代表大会及其常委会作为最高国家权利机关，在<pro>(true)建设社会主义法制国家的历史进程中担负着<pro>(true)制定法律和监督法律实施的重要职责。 \\
    Qwen3-8b & false false \\
    & \textbf{Error Analysis}: False Negative. Fails to link institutional NPs with subsequent ZPs in formal language. \\
    DeepSeek & true true \\
    & \textbf{Error Analysis}: Correct prediction. \\
    \midrule
    Example 2 & <pro>(false)别到时候整整错了,<pro>(true)在不是免费的. \\
    Qwen3-8b & true true \\
    & \textbf{Error Analysis}: False Positive. Influenced by colloquial patterns, defaults to assuming explicit antecedents. \\
    DeepSeek & true true \\
    & \textbf{Error Analysis}: False Positive. Over-reliant on surface syntax, misinterpreting generic omissions. \\

    \midrule
    \textbf{Task 3} & \textbf{Referential Type Classification} \\
    \midrule
    Example 1 & 但其广告牌是另一件事情．洛杉矶居民要求<pro>(Cataphora)除去这些牌子及其内容．而制片厂正在让步. \\
    Qwen3-8b & Deixis \\
    & \textbf{Error Analysis}: Misclassification. Fails to identify cataphoric relationship, defaulting to deictic reference. \\
    DeepSeek & Cataphora \\
    & \textbf{Error Analysis}: Correct prediction. \\
    \midrule
    Example 2 & 在东海村核灾事故当中３名现场作业员以钢桶将多达１６公斤的过量油燃料注入沉淀槽而引起了临界反应，３个人遭到大量幅射线的照射，有２个人在医院丧生，唯一生还者恒川后来出院回家，这起事故造成了至少４３９人遭到幅射照射，并且迫使临近地区的３２万名原著民躲在家中避难长达１天以上，目前东海村油燃料加工厂仍然维持<pro>(Anaphora)关闭的状态。\\
    Qwen3-8b & Deixis \\
    & \textbf{Error Analysis}: Misclassification. Ignores discourse context, failing to resolve the anaphora. \\
    DeepSeek & Anaphora \\
    & \textbf{Error Analysis}: Correct prediction. \\

    \midrule
    \textbf{Task 4} & \textbf{ZP Resolution} \\
    \midrule
    Example 1 & 嗯,嗯,没有东西叫妈妈.没有,<pro>没有寄,没有寄.什么毛线衣啊,什么东西的. \\
    Qwen3-8b & \textbf{[MCQ]:} pred=['东西', '妈妈', '毛线衣'] gold=['妈妈', '*pro*'] result=RIGHT \\
     & \textbf{[QA]:} pred=['妈妈', '东西'] gold=['妈妈', '*pro*'] result=RIGHT \\
     & \textbf{[Clustering]:} 嗯,嗯,没有<东西>\{\#0\}叫<妈妈>\{\#1\}.没有,<pro>\{\#1\}没有寄,没有寄.什么<毛线衣>\{\#2\}啊,什么<东西>\{\#0\}的. pred={(9, '妈'), (10, '妈')} gold={(9, '妈'), (10, '妈')} correct=2 \\
    & \textbf{Error Analysis}: Successful resolution. Qwen3-8b effectively maintains the discourse entity track, correctly linking the zero pronoun to the specific entity "妈妈" using multi-modal information. \\
    DeepSeek & \textbf{[MCQ]:} pred=['东西'] gold=['妈妈', '*pro*'] result=WRONG \\
     & \textbf{[QA]:} pred=['东西'] gold=['妈妈', '*pro*'] result=WRONG \\
     & \textbf{[Clustering]:} 嗯,嗯,没有<东西>\{\#0\}叫<妈妈>\{\#1\}.没有,<pro>\{\#0\}没有寄,没有寄.什么<毛线衣>\{\#2\}啊,什么<东西>\{\#0\}的. pred={(6, '东'), (33, '西'), (7, '西'), (32, '东')} gold={(9, '妈'), (10, '妈')} correct=0 \\
    & \textbf{Error Analysis}: Failure in antecedent selection. DeepSeek exhibits a "frequency bias," incorrectly resolving the zero pronoun to the high-frequency entity "东西" appearing earlier in the clause, while failing to capture the pragmatic intent of the speaker. \\
    \midrule
    Example 2 & 受不了饥饿的痛苦时，塞勒斯会吃大量的食物，然后呕吐以保持体重．他说<pro>去骑师更衣室走一圈你就知道这法子多简单且已被人接受．这些是正规的马桶。 \\
    Qwen3-8b & \textbf{[MCQ]:} pred=['塞勒斯', '法子'] gold=['*pro*', '你'] result=WRONG \\
     & \textbf{[QA]:} pred=['骑师更衣室', '正规的马桶'] gold=['*pro*', '你'] result=WRONG \\
     & \textbf{[Clustering]:} 受不了<饥饿>\{\#0\}的<痛苦>\{\#0\}时，<塞勒斯>\{\#1\}会吃大量的<食物>\{\#2\}，然后呕吐以保持<体重>\{\#3\}．他说<pro>\{\#1\}去<骑师>\{\#4\}<更衣室>\{\#5\}走一圈你就知道这<法子>\{\#6\}多简单且已被<人>\{\#7\}接受．这些是正规的<马桶>\{\#8\}。 pred={(12, '斯'), (10, '塞'), (11, '勒')} gold={(42, '你')} correct=0 \\
    & \textbf{Error Analysis}: Failure to model participant shift. The model treats the zero pronoun as referring to the preceding subject "塞勒斯," failing to recognize the transition to an imperative structure that implies a second-person reference ("你").\\
    DeepSeek & \textbf{[MCQ]:} pred=['塞勒斯'] gold=['*pro*', '你'] result=WRONG \\
    & \textbf{[QA]:} pred=['塞勒斯'] gold=['*pro*', '你'] result=WRONG \\
    & \textbf{[Clustering]:} 受不了<饥饿>\{\#0\}的<痛苦>\{\#1\}时，<塞勒斯>\{\#2\}会吃大量的<食物>\{\#3\}，然后呕吐以保持<体重>\{\#4\}．他说<pro>\{\#2\}去<骑师>\{\#5\}<更衣室>\{\#6\}走一圈你就知道这<法子>\{\#7\}多简单且已被<人>\{\#8\}接受．这些是正规的<马桶>\{\#9\}． pred={(11, '勒'), (12, '斯'), (10, '塞')} gold={(42, '你')} correct=0 \\
    & \textbf{Error Analysis}: Semantic over-reliance on local discourse subjects. The model fails to account for the quoted speech context, leading to a "subject-persistence" error where it incorrectly re-assigns the action to the main subject of the previous clause.\\
    \bottomrule
\caption{Error analysis of LLMs (Qwen3-8B and DeepSeek-V3.2) on Chinese Zero Pronoun tasks.} \label{tab:error_analysis} \\
\end{longtable}}

%% file: tab/prompt_templates.tex
\onecolumn 
\small
\begin{longtable}{p{15cm}}
    \toprule
    \textbf{Identification} \\
    \midrule
    \# 任务说明：零代词指在语篇中本应出现（通常是主语或宾语等）的代词却被省略，但从语义或句法结构上看，该代词对表意是必需的。即使无法准确推断出“谁”或“什么”被省略，只要存在必要成分的省略，就视为零代词。请你阅读给定的目标中文文本，按照指定规范对其中的零代词缺失位置进行标注，不用展示你的思考过程。
\# 任务要求\\
1. 标注方式：仅在文本中插入<pro>，不得添加、删除或修改任何除<pro>以外的多余内容，保持原句的换行、空格、顺序、标点和其他字符不变。\\
2. 检测范围：只分析和标注标准目标中文文本实例中的中文部分，忽略其中出现的英文、数字、符号或无意义空白等内容。\\
3. 判定标准：判断该分句是否省略了必需的主语或宾语等成分。若该成分缺失但原本在语义或句法结构上应当出现，则确认为零代词。即使无法从上下文中明确推断出“谁”或“什么”，也需要在省略的地方标注零代词。\\
4. 输出格式：最终输出只包含标注后的目标文本，不包含任何解释、分析或多余说明。同时保持与原始文本相同的行、段落结构，只在零代词缺失处插入<pro>标记。\\
\# 示例\\
以下展示三个常见的零代词场景作为你标注时的重要参考：\\
1. 省略主语\\
原句：“张三是一名学生，学习很努力。”\\
标注后：“张三是一名学生，<pro>学习很努力。”（第二个分句缺少对“张三”的指代。）\\
2. 省略主语（不明确）\\
原句：“明天去爬山吗？”\\
标注后：“<pro>明天去爬山吗？”（句首省略了主语，即使不确定到底是“你”“我们”或其他，也要插入<pro>。）\\
3. 省略宾语\\
原句：“我吃完了。”（若为完整表达，但想突出“吃完了什么”则需要信息判断；如果明确有缺省内容则标注）\\
如果上文语境明确省略了“苹果”等代词位置，则可在动词后插入<pro>。但请只在真正存在零代词缺失的地方插入<pro>，不要在所有谓语动词后都胡乱插入。\\
\# 操作步骤\\
1. 分句。先将目标文本拆分为主要分句（包括主句、从句、副句、并列句等）。\\
2. 判断省略。对于每个分句，判断该分句的主要谓语动词是否存在主语、宾语等必要成分的缺省。\\
3. 插入<pro>。如果确认省略了必要的代词，则在此处插入<pro>。\\
4. 保持结构。除插入<pro>外，不要改动或删除其他内容。\\
5. 输出。只返回标注完成的文本，不包含任何分析或说明。\\
请在确保了解以上信息后阅读下面目标中文文本，在每一个零代词缺失的位置插入<pro>：\\
    \midrule
    \textbf{Referentiality Classification} \\
    \midrule
   \# \textbf{任务说明}：零代词指在语篇中本应出现（通常是主语或宾语等）的代词却被省略，但从语义或句法结构上看，该代词对表意是必需的。即使无法准确推断出“谁”或“什么”被省略，只要存在必要成分的省略，就视为零代词。请逐行阅读目标中文文本，判断文中给出的一个零代词（文本中用<pro>标记）是否可指，若是则返回T，否则返回F，不用展示你的思考过程。\\
\# 任务要求：\\
1. 划定检测范围：仅分析目标文本中的汉语部分，忽略英文、数字、符号等无关内容。\\
2. 严格遵循返回格式：只需返回判断一个字母（T或者F）表示判断结果，无需添加其他任何分析内容。\\
3. 明确零代词定义：句子中语义或句法结构上必需的代词（主语或宾语等）被省略，但该成分在语义或句法上应当出现，则为零代词。\\
4. 明确零代词可指性定义：若目标零代词在所给上下文语境中有实际指代的内容，那么无论该指代内容在文中是否显式出现，都称该零代词可指。\\
\# 示例\\
以下展示零代词的可指性作为你标注时的重要参考：\\
1.可指\\
(I) 前指：\\
原句：“张三是一名学生，<pro>学习很努力。”\\
结论：T\\
分析：第二个分句缺少对“张三”的指代，而“张三”作为先行词显性地出现在零代词之前，故<pro>可指。\\
(II) 后指：\\
原句：“<pro>明天去爬山吗？我们明天都没课诶！”\\
结论：T\\
分析：句首零代词<pro>指代出现在第二个分句中的“我们”，即<pro>的指代内容显性地出现了，故<pro>可指。\\
(III) 指别：\\
原句：“如果<pro>不是的话<pro>全部都不可以进入，尤其是<pro>不可以摄录机拍摄。”\\
结论：T\\
分析：两个分句零代词省略的主语都属于一个共指链，指代内容相同，属于deixis，故<pro>可指。\\
2.不可指\\
原句：“现在<pro>距离香港迪斯尼乐园九月十二号的开业只有一个月的时间了，通往迪斯尼的地铁也已经建好。”\\
结论：F\\
分析：该零代词成分在语义上应当存在，但其既不属于某个共指链也无法通过上下文判断被省略的是“谁”或“什么”，即该零代词没有实际的指代内容，故<pro>不可指。\\
请在确保了解以上信息后判断以下目标中文文本中的零代词<pro>是否可指，仅返回T或者F：\\
    \midrule
    \textbf{Referential Type Classification} \\
    \midrule
    \# 任务说明：零代词指在语篇中本应出现（通常是主语或宾语等）的代词却被省略，但从语义或句法结构上看，该代词对表意是必需的。即使无法准确推断出“谁”或“什么”被省略，只要存在必要成分的省略，就视为零代词。请逐行阅读目标文本，判断文中给出的一个零代词（文本中用<pro>标记）的指代类别是以下选项的哪一个？请直接返回你判断的选项字母，不要推理。\\
\# 任务要求\\
1. 你的输出必须且只能是A、B、C三个字符之一，严禁在你的输出中复述题目、分析推理、添加标点或者换行。\\
2. 划定检测范围：仅分析目标文本中的汉语以判断零代词<pro>的指代类型，忽略英文、数字、符号等无关内容。\\
\# 示例\\
以下展示零代词各指代类型的示例作为你标注时的重要参考：\\
\quad A 前指：指某一语言成分的语义依赖于前置表达的表达方式，指代内容为上文中已出现过的另一语言成分。\\
\quad \quad输入：张三是一名学生，<pro>学习很努力。\\
\quad \quad输出: A \\
\quad B 后指：指某一语言成分的语义依赖于后置表达的表达方式，预先指代下文中即将出现的另一语言成分。\\
\quad \quad输入：<pro>明天去爬山吗？我们明天都没课诶！\\
\quad \quad输出: B\\
\quad C 指别：指描述词语或表达必须依赖具体语境或说话者、时间、地点、手势等现实情境才能确定其含义的现象。\\
\quad \quad输入：今天天气真好，<pro>可以在公园野餐放松放松。\\
\quad \quad输出: C \\
\# 输出格式：请以"A" "B" "C"的格式回答，不要添加其他内容。\\
\# 目标文本：\\
    \midrule

    \textbf{MCQ-based Resolution} \\
    \midrule
    \# 任务说明：零代词指在语篇中本应出现（通常是主语或宾语等）的代词却被省略，但从语义或句法结构上看，该代词对表意是必需的。即使无法准确推断出“谁”或“什么”被省略，只要存在必要成分的省略，就视为零代词。请仔细阅读下方文段并判断文中零代词<pro>指向哪些候选项。\\
\# 任务要求\\
1. 输出仅为候选项字母（大写），若有多个，用单个空格分隔；例如：A B C。不允许输出任何解释、标点、注释或多余文字。\\
2. 若判断有多个可能先行词，请把所有可能项列出。 \\
\# 示例\\
文段：小明去了商店，<pro>买了很多水果。\\
选项：A. 小明 B. 商店 C. 水果 \\ 
输出：A\\
\# 文段：\\
\# 选项：\\
\# 输出：\\
    \midrule

    \textbf{QA-based Resolution} \\
    \midrule
    \# 任务说明：零代词指在语篇中本应出现（通常是主语或宾语等）的代词却被省略，但从语义或句法结构上看，该代词对表意是必需的。即使无法准确推断出“谁”或“什么”被省略，只要存在必要成分的省略，就视为零代词。请阅读下方文段，识别出文中零代词<pro>指代的名词短语（名词短语指示性成分，如“张三”“那辆车”“美国总统”等）。\\
\# 任务要求\\
1. 输出仅包含所有可能的名词短语，若有多个答案，用单个空格分隔；不允许输出序号、标签、解释或任何多余文字。\\
2. 若出现指代模糊且多个短语均合理，则列出所有合理短语；尽量按在原文中出现的先后顺序给出。\\
3. 名词短语应与原词一致（不进行同义替换或改写）。\\
\# 输出格式\\
只返回词，不要返回括号、引号或注释。正确输出示例：张三 李四（多个答案用空格分隔）\\
\# 示例\\  
1. 单一先行词 \\ 
\quad 文段：张三去了学校，<pro>在教室等他。\\  
\quad 输出：张三\\
2. 多个可能先行词：\\
\quad 文段：张三和李四来了，<pro>都很高兴。\\  
\quad 输出：张三 李四 \\ 
\# 文段：\\
\# 输出：\\
    \midrule

    \textbf{Clustering-based Resolution} \\
    \midrule
    \# 任务说明：零代词指在语篇中本应出现（通常是主语或宾语等）的代词却被省略,但从语义或句法结构上看,该代词对表意是必需的。即使无法准确推断出“谁”或“什么”被省略,只要存在必要成分的省略,就视为零代词。请对输入文段进行共指链标注,包括所有实体与零代词。\\
\# 任务要求\\
1. 使用Markdown共指格式：$<$实体$>${\#cluster\_id}。\\
2. cluster\_id从0开始连续编号，相同指代实体使用相同编号。\\
3. 文段中出现的<pro>都必须被视为一个实体,并必须在其后加上{\#cluster\_id},任何情况下都不允许遗漏、跳过、忽略<pro>的标注。\\
4. 只用输出标注后的文本,输出必须保持原文顺序,不得删词、改词、增词。\\
\# 示例\\
原文：<<张三><李四>>去了学校,<pro>见到了<王五>。\\
标注：<<张三>{\#0}<李四>{\#1}>{\#2}去了学校,<pro>{\#2}见到了<王五>{\#3}。\\
\# 文段：\\
\# 输出：\\
    \midrule

    \textbf{ZP-Unaware Translation} \\
    \midrule
    \# 任务说明：请将给出的文本直接翻译成英文。输入为一行中文，输出仅一行译文即可，不添加任何注释说明。\\
\# 待翻译文本：\\
    \midrule

    \textbf{ZP-Aware Translation} \\
    \midrule
    \# 任务说明：零代词指在语篇中本应出现（通常是主语或宾语等）的代词却被省略，但从语义或句法结构上看，该代词对表意是必需的。即使无法准确推断出“谁”或“什么”被省略，只要存在必要成分的省略，就视为零代词。请将一行包含中文零代词的文本翻译成英文。\\
\# 任务要求\\
1. 译文在词级上需能和原文词对译，句级上需保留完整结构和原有语序，语义上需完整体现中文原文中的所有显性与隐性信息不得有任何省略。\\
2. 为追求译文的精确，翻译之前通过语法分析潜在零代词的位置、可指性、类型与指代内容是被鼓励的。\\
3. 若被找出的零代词不可指，则不需要强行译出。\\
\# 输出格式：输出一行英文译文，不包含任何解释、注释或中间推理。\\
\# 示例\\
中文输入：张三是一名学生，学习很努力。\\
\#\# 步骤：\\
1. 通过寻找发现“学习”一词前缺少主语，存在零代词<pro>。\\
2. 联系上下文，判断出<pro>为可指零代词且指代内容即为上文的“张三”。\\
3. 恢复零代词主语的指代内容"he"或者"Zhang San"，并在译文中显式表达如下：Zhang San is a student and he studies very hard.\\
\# 待翻译文本：\\

    \midrule

    \textbf{Oracle Translation} \\
    \midrule
    \# 任务说明：零代词指在语篇中本应出现（通常是主语或宾语等）的代词却被省略，但从语义或句法结构上看，该代词对表意是必需的。在输入文本中，标记<pro>仅作为零代词占位符，表示该位置存在一个被省略、但必须在英文中显式表达的指代成分。你的任务是，将一行包含<pro>标记的中文文本翻译成英文，并在译文中用合适的英文代词或名词短语替换每一个<pro>。\\
\# 任务要求\\
1. 每一个<pro>必须在英文译文中被显式替换为具体的指代形式，严禁保留、照抄或省略<pro>。\\
2. 若指代对象明确，使用文中的人称代词或名词进行翻译；若无法唯一确定指代对象，必须使用语义合理的泛指表达。\\
3. 译文需在词级和句级结构上与原文保持对应，完整表达原文中所有显性与隐性信息。\\
4. 输出前请进行自检，若译文中仍出现<pro>，请重新生成译文。\\
\# 输出格式 \\
仅输出一行英文译文。不得包含<pro>、中文、解释、注释或中间推理。\\
\# 示例 \\
中文输入：张三是一名学生，<pro>学习很努力。\\
英文输出：Zhang San is a student, he studies very hard.\\
\# 待翻译文本：\\
    \bottomrule
    \caption{Prompt templates for all evaluation tasks related to Zero Pronoun processing.} 
    \label{tab:prompt_templates}
\end{longtable}
\twocolumn

%% file: acl_latex.bbl
\begin{thebibliography}{31}
\providecommand{\natexlab}[1]{#1}

\bibitem[{Bagga(1998)}]{bagga1998algorithms}
Amit Bagga. 1998.
\newblock Algorithms for scoring coreference chains.
\newblock In \emph{Proc. Linguistic Coreference Workshop at the first Conf. on Language Resources and Evaluation (LREC), Granada, Spain, May 1998}.

\bibitem[{Balloccu et~al.(2024)Balloccu, Schmidtov{\'a}, Lango, and Dusek}]{balloccu-etal-2024-leak}
Simone Balloccu, Patr{\'i}cia Schmidtov{\'a}, Mateusz Lango, and Ondrej Dusek. 2024.
\newblock \href {https://doi.org/10.18653/v1/2024.eacl-long.5} {Leak, cheat, repeat: Data contamination and evaluation malpractices in closed-source {LLM}s}.
\newblock In \emph{Proceedings of the 18th Conference of the European Chapter of the Association for Computational Linguistics (Volume 1: Long Papers)}, pages 67--93, St. Julian{'}s, Malta. Association for Computational Linguistics.

\bibitem[{Cao(1979)}]{cao1979functional}
Fengfu Cao. 1979.
\newblock \emph{A functional study of topic in Chinese: The first step towards discourse analysis}, volume~3.
\newblock Student Book Co.

\bibitem[{Cardie and Wagstaff(1999)}]{cardie-wagstaff-1999-noun}
Claire Cardie and Kiri Wagstaff. 1999.
\newblock \href {https://aclanthology.org/W99-0611/} {Noun phrase coreference as clustering}.
\newblock In \emph{1999 Joint {SIGDAT} Conference on Empirical Methods in Natural Language Processing and Very Large Corpora}.

\bibitem[{Chen and Ng(2013)}]{chen-ng-2013-chinese}
Chen Chen and Vincent Ng. 2013.
\newblock \href {https://aclanthology.org/D13-1135/} {{C}hinese zero pronoun resolution: Some recent advances}.
\newblock In \emph{Proceedings of the 2013 Conference on Empirical Methods in Natural Language Processing}, pages 1360--1365, Seattle, Washington, USA. Association for Computational Linguistics.

\bibitem[{Chen and Ng(2014)}]{chen2014chinese}
Chen Chen and Vincent Ng. 2014.
\newblock Chinese overt pronoun resolution: A bilingual approach.
\newblock In \emph{Proceedings of the AAAI Conference on Artificial Intelligence}, volume~28.

\bibitem[{Chen(2022)}]{chen2022computational}
Guanyi Chen. 2022.
\newblock \emph{Computational generation of Chinese noun phrases}.
\newblock Ph.D. thesis, Utrecht University.

\bibitem[{Chen et~al.(2018)Chen, van Deemter, and Lin}]{chen-etal-2018-modelling}
Guanyi Chen, Kees van Deemter, and Chenghua Lin. 2018.
\newblock \href {https://doi.org/10.18653/v1/W18-6519} {Modelling pro-drop with the rational speech acts model}.
\newblock In \emph{Proceedings of the 11th International Conference on Natural Language Generation}, pages 159--164, Tilburg University, The Netherlands. Association for Computational Linguistics.

\bibitem[{Guo et~al.(2025)Guo, Yang, Zhang, Song, Zhang, Xu, Zhu, Ma, Wang, Bi et~al.}]{guo2025deepseek}
Daya Guo, Dejian Yang, Haowei Zhang, Junxiao Song, Ruoyu Zhang, Runxin Xu, Qihao Zhu, Shirong Ma, Peiyi Wang, Xiao Bi, and 1 others. 2025.
\newblock Deepseek-r1: Incentivizing reasoning capability in llms via reinforcement learning.
\newblock \emph{arXiv preprint arXiv:2501.12948}.

\bibitem[{Huang(1984)}]{huang1984distribution}
C-T~James Huang. 1984.
\newblock On the distribution and reference of empty pronouns.
\newblock \emph{Linguistic inquiry}, pages 531--574.

\bibitem[{Huang(1989)}]{huang1989pro}
C-T~James Huang. 1989.
\newblock Pro-drop in chinese: A generalized control theory.
\newblock In \emph{The null subject parameter}, pages 185--214. Springer.

\bibitem[{Huang and Liu(2024)}]{huang2024evaluating}
Yan Huang and Wei Liu. 2024.
\newblock Evaluating the translation performance of large language models based on euas-20.
\newblock \emph{arXiv preprint arXiv:2408.03119}.

\bibitem[{Hurst et~al.(2024)Hurst, Lerer, Goucher, Perelman, Ramesh, Clark, Ostrow, Welihinda, Hayes, Radford et~al.}]{hurst2024gpt}
Aaron Hurst, Adam Lerer, Adam~P Goucher, Adam Perelman, Aditya Ramesh, Aidan Clark, AJ~Ostrow, Akila Welihinda, Alan Hayes, Alec Radford, and 1 others. 2024.
\newblock Gpt-4o system card.
\newblock \emph{arXiv preprint arXiv:2410.21276}.

\bibitem[{Jones and Bergen(2025)}]{jones2025large}
Cameron~R Jones and Benjamin~K Bergen. 2025.
\newblock Large language models pass the turing test.
\newblock \emph{arXiv preprint arXiv:2503.23674}.

\bibitem[{Konno et~al.(2021)Konno, Kiyono, Matsubayashi, Ouchi, and Inui}]{konno-etal-2021-pseudo}
Ryuto Konno, Shun Kiyono, Yuichiroh Matsubayashi, Hiroki Ouchi, and Kentaro Inui. 2021.
\newblock \href {https://doi.org/10.18653/v1/2021.emnlp-main.308} {Pseudo zero pronoun resolution improves zero anaphora resolution}.
\newblock In \emph{Proceedings of the 2021 Conference on Empirical Methods in Natural Language Processing}, pages 3790--3806, Online and Punta Cana, Dominican Republic. Association for Computational Linguistics.

\bibitem[{Le and Ritter(2023)}]{le2023large}
Nghia~T Le and Alan Ritter. 2023.
\newblock Are large language models robust coreference resolvers?
\newblock \emph{arXiv preprint arXiv:2305.14489}.

\bibitem[{Liu et~al.(2025)Liu, Mei, Lin, Xue, Wang, Xu, Wu, Zhang, Lin, Dong et~al.}]{liu2025deepseek}
Aixin Liu, Aoxue Mei, Bangcai Lin, Bing Xue, Bingxuan Wang, Bingzheng Xu, Bochao Wu, Bowei Zhang, Chaofan Lin, Chen Dong, and 1 others. 2025.
\newblock Deepseek-v3. 2: Pushing the frontier of open large language models.
\newblock \emph{arXiv preprint arXiv:2512.02556}.

\bibitem[{Liu et~al.(2024)Liu, Shen, Zhu, Xu, Qian, Song, Zhang, Tang, Zhang, Yang et~al.}]{liu2024systematic}
Yikang Liu, Yeting Shen, Hongao Zhu, Lilong Xu, Zhiheng Qian, Siyuan Song, Kejia Zhang, Jialong Tang, Pei Zhang, Baosong Yang, and 1 others. 2024.
\newblock A systematic assessment of language models with linguistic minimal pairs in chinese.
\newblock \emph{arXiv preprint arXiv:2411.06096}.

\bibitem[{Mao et~al.(2024{\natexlab{a}})Mao, Chen, Zhang, Guerin, and Cambria}]{mao-etal-2024-gpteval}
Rui Mao, Guanyi Chen, Xulang Zhang, Frank Guerin, and Erik Cambria. 2024{\natexlab{a}}.
\newblock \href {https://aclanthology.org/2024.lrec-main.693/} {{GPTE}val: A survey on assessments of {C}hat{GPT} and {GPT}-4}.
\newblock In \emph{Proceedings of the 2024 Joint International Conference on Computational Linguistics, Language Resources and Evaluation (LREC-COLING 2024)}, pages 7844--7866, Torino, Italia. ELRA and ICCL.

\bibitem[{Mao et~al.(2024{\natexlab{b}})Mao, He, Zhang, Chen, Ni, Yang, and Cambria}]{mao2024survey}
Rui Mao, Kai He, Xulang Zhang, Guanyi Chen, Jinjie Ni, Zonglin Yang, and Erik Cambria. 2024{\natexlab{b}}.
\newblock A survey on semantic processing techniques.
\newblock \emph{Information Fusion}, 101:101988.

\bibitem[{Song et~al.(2020)Song, Xu, Zhang, Chen, and Yu}]{song-etal-2020-zpr2}
Linfeng Song, Kun Xu, Yue Zhang, Jianshu Chen, and Dong Yu. 2020.
\newblock \href {https://doi.org/10.18653/v1/2020.acl-main.482} {{ZPR}2: Joint zero pronoun recovery and resolution using multi-task learning and {BERT}}.
\newblock In \emph{Proceedings of the 58th Annual Meeting of the Association for Computational Linguistics}, pages 5429--5434, Online. Association for Computational Linguistics.

\bibitem[{Vilain et~al.(1995)Vilain, Burger, Aberdeen, Connolly, and Hirschman}]{vilain-etal-1995-model}
Marc Vilain, John Burger, John Aberdeen, Dennis Connolly, and Lynette Hirschman. 1995.
\newblock \href {https://aclanthology.org/M95-1005/} {A model-theoretic coreference scoring scheme}.
\newblock In \emph{Sixth Message Understanding Conference ({MUC}-6): Proceedings of a Conference Held in {C}olumbia, {M}aryland, November 6-8, 1995}.

\bibitem[{Wang et~al.(2023)Wang, Liu, Xu, Song, Shi, and Tu}]{wang-etal-2023-survey}
Longyue Wang, Siyou Liu, Mingzhou Xu, Linfeng Song, Shuming Shi, and Zhaopeng Tu. 2023.
\newblock \href {https://doi.org/10.18653/v1/2023.acl-long.187} {A survey on zero pronoun translation}.
\newblock In \emph{Proceedings of the 61st Annual Meeting of the Association for Computational Linguistics (Volume 1: Long Papers)}, pages 3325--3339, Toronto, Canada. Association for Computational Linguistics.

\bibitem[{Wang et~al.(2018)Wang, Tu, Shi, Zhang, Graham, and Liu}]{wang2018translating}
Longyue Wang, Zhaopeng Tu, Shuming Shi, Tong Zhang, Yvette Graham, and Qun Liu. 2018.
\newblock Translating pro-drop languages with reconstruction models.
\newblock \emph{arXiv preprint arXiv:1801.03257}.

\bibitem[{Wang et~al.(2019)Wang, Tu, Wang, and Shi}]{wang-etal-2019-one}
Longyue Wang, Zhaopeng Tu, Xing Wang, and Shuming Shi. 2019.
\newblock \href {https://doi.org/10.18653/v1/D19-1085} {One model to learn both: Zero pronoun prediction and translation}.
\newblock In \emph{Proceedings of the 2019 Conference on Empirical Methods in Natural Language Processing and the 9th International Joint Conference on Natural Language Processing (EMNLP-IJCNLP)}, pages 921--930, Hong Kong, China. Association for Computational Linguistics.

\bibitem[{Xu et~al.(2022)Xu, Wang, Wong, Liu, Song, Chao, Shi, and Tu}]{xu-etal-2022-guofeng}
Mingzhou Xu, Longyue Wang, Derek~F. Wong, Hongye Liu, Linfeng Song, Lidia~S. Chao, Shuming Shi, and Zhaopeng Tu. 2022.
\newblock \href {https://doi.org/10.18653/v1/2022.emnlp-main.774} {{G}uo{F}eng: A benchmark for zero pronoun recovery and translation}.
\newblock In \emph{Proceedings of the 2022 Conference on Empirical Methods in Natural Language Processing}, pages 11266--11278, Abu Dhabi, United Arab Emirates. Association for Computational Linguistics.

\bibitem[{Yang et~al.(2025)Yang, Li, Yang, Zhang, Hui, Zheng, Yu, Gao, Huang, Lv et~al.}]{yang2025qwen3}
An~Yang, Anfeng Li, Baosong Yang, Beichen Zhang, Binyuan Hui, Bo~Zheng, Bowen Yu, Chang Gao, Chengen Huang, Chenxu Lv, and 1 others. 2025.
\newblock Qwen3 technical report.
\newblock \emph{arXiv preprint arXiv:2505.09388}.

\bibitem[{Yin et~al.(2018{\natexlab{a}})Yin, Zhang, Zhang, Liu, and Wang}]{yin-etal-2018-deep}
Qingyu Yin, Yu~Zhang, Wei-Nan Zhang, Ting Liu, and William~Yang Wang. 2018{\natexlab{a}}.
\newblock \href {https://doi.org/10.18653/v1/P18-1053} {Deep reinforcement learning for {C}hinese zero pronoun resolution}.
\newblock In \emph{Proceedings of the 56th Annual Meeting of the Association for Computational Linguistics (Volume 1: Long Papers)}, pages 569--578, Melbourne, Australia. Association for Computational Linguistics.

\bibitem[{Yin et~al.(2017)Yin, Zhang, Zhang, and Liu}]{yin-etal-2017-chinese}
Qingyu Yin, Yu~Zhang, Weinan Zhang, and Ting Liu. 2017.
\newblock \href {https://doi.org/10.18653/v1/D17-1135} {{C}hinese zero pronoun resolution with deep memory network}.
\newblock In \emph{Proceedings of the 2017 Conference on Empirical Methods in Natural Language Processing}, pages 1309--1318, Copenhagen, Denmark. Association for Computational Linguistics.

\bibitem[{Yin et~al.(2018{\natexlab{b}})Yin, Zhang, Zhang, Liu, and Wang}]{yin-etal-2018-zero}
Qingyu Yin, Yu~Zhang, Weinan Zhang, Ting Liu, and William~Yang Wang. 2018{\natexlab{b}}.
\newblock \href {https://aclanthology.org/C18-1002/} {Zero pronoun resolution with attention-based neural network}.
\newblock In \emph{Proceedings of the 27th International Conference on Computational Linguistics}, pages 13--23, Santa Fe, New Mexico, USA. Association for Computational Linguistics.

\bibitem[{Zhang et~al.(2019)Zhang, Liu, Yin, and Zhang}]{zhang2019neural}
Weinan Zhang, Ting Liu, Qingyu Yin, and Yu~Zhang. 2019.
\newblock Neural recovery machine for chinese dropped pronoun.
\newblock \emph{Frontiers of Computer Science}, 13(5):1023--1033.

\end{thebibliography}
